\documentclass{article}
\usepackage{graphicx}
\usepackage{graphics}
\usepackage{amsfonts}
\usepackage{marvosym}

\begin{document}

\title{Ultrametric Wavelet Regression of Multivariate Time Series:
Application to Colombian Conflict Analysis}
\author{Fionn Murtagh \\
Science Foundation Ireland, Wilton Place, Dublin 2, Ireland, and \\
Department of Computer Science, Royal Holloway, University of London \\
Egham TW20 0EX, England \\
Email: fmurtagh@acm.org \\
\\
Michael Spagat \\
Department of Economics, Royal Holloway, University of London \\
Egham TW20 0EX, England \\
Email: m.spagat@rhul.ac.uk \\
\\
Jorge A. Restrepo \\
Conflict Analysis Resource Center, Carrera 10 No 65-35 Of. 703 \\
Bogot\'a, Colombia \\
Email: jorge.restrepo@cerac.org.co }
\maketitle

\begin{abstract}
We first pursue the study of how hierarchy provides a well-adapted tool for
the analysis of change. Then, using a time sequence-constrained hierarchical
clustering, we develop the practical aspects of a new approach to wavelet
regression. This provides a new way to link hierarchical relationships in a
multivariate time series data set with external signals. Violence data from
the Colombian conflict in the years 1990 to 2004 is used throughout. We
conclude with some proposals for further study on the relationship between
social violence and market forces, viz.\ between the Colombian conflict 
and the US narcotics market.
\end{abstract}


\section{Introduction}

\label{sect1}

In this article we will refer to Appendices 2, 3 and 4 for a basic
introduction to Correspondence Analysis, sequence-constrained hierarchical
clustering and the Haar wavelet transform on a hierarchy. In section 
\ref{sect11} we will first review the terminology that we use.

\subsection{Terminology: Ultrametric and Wavelet Regression}

\label{sect11}

By ultrametric we intend a precise definition of ``hierarchy'' which as such
is a loose concept. The ultrametric inequality holds for pairwise distances
(for a distance, $d$, defined for all pairs of points $i, j$ as a symmetric,
positive-definite mapping to the reals, the ultrametric inequality holds for
all triplets of points, $i, j, k$: $d(i,k) \leq \mbox{max} \{ d(i,j), d(j,k)
\}$). A set of ultrametric distances can be derived from a binary (i.e.,
each non-terminal node has either one or two child nodes; a terminal node
has no child node), ranked (i.e., each node in the hierarchy has a ``level''
value associated which is typically positive real-valued) rooted tree 
\cite{benz}. Reciprocally, a binary, ranked rooted tree, known as a dendrogram,
is a representation of an ultrametric set of distances. We can relax the
binary property of this tree but often it is algorithmically convenient to
impose this.

By regression, we mean some mapping of independent or explanatory variable $x
$ onto a dependent or response variable, $y$, such that the behavior of $y$
as a function of $x$ can be studied. In this work, $x$ is multidimensional,
while $y$ is unidimensional. A novelty is that $x$ is ultrametric, i.e.\ all
triplets of points satisfy the ultrametric inequality; or we avail of a
hierarchical structure on $x$.

Wavelet regression or wavelet smoothing is ``potentially highly locally
adaptive'' \cite{nason}. A data signal or set of points is projected into a
new basis such that spatial and frequency properties are brought out or
exposed. (Consider, for comparative purposes, the following. A sinusoidal
basis system used in Fourier analysis helps with frequency properties but
not with spatial. A basis of principal components helps with spatial
properties but not with frequency properties.) The wavelet transform
projection itself is invertible. This means that data can be reconstructed,
after carrying out modifications in transform space. (The same principle can
be applied in Fourier analysis or principal component analysis.)
Modification in wavelet space typically involves setting low-valued wavelet
coefficients to zero, leading to a smoother reconstruction. A novelty in our
work is that we understand wavelet regression to involve both explanatory
and response variables, rather than just wavelet smoothing of erstwhile
explanatory variables.

As we will see below, our wavelet transform is a hierarchical one, in the
sense of being carried out on a hierarchical clustering or dendrogram. It is
therefore a wavelet transform in an ultrametric space, or an ultrametric
wavelet transform.

In this context, the effect of wavelet coefficient thresholding leads to
piecewise constant function approximation. We will sketch out the
explanation for this. The subnodes of any given node are determined in
transform space by a positive or a negative detail vector. (The
dimensionality of this vector is the dimensionality of the points we are
working on. It is possible that these points are single valued.) So left and
right subnodes are defined as the parent node's vector, plus or minus
(respectively) the detail vector. Setting the detail vector to zero just
means that the subnodes, in the reconstructed or wavelet-smoothed data, now
become, both identically, equal to the parent node. In other words, within
the cluster defined by the parent node, we now have a (constant within the
cluster) vector given by the parent node. Wavelet smoothing, in the way that
we are implementing it here, furnishes a constant vector within a cluster.

\subsection{Time Series Segmentation and the Issue of Normalization}

The approach we develop in this article is based on the following.

\begin{enumerate}
\item A goal is ``subsequence clustering'' or segmentation.

\item Normalization and weighting of time series and attributes are
addressed.

\item We closely interface the analysis with interpretation. We regress the
hierarchical segmentation on external time series.
\end{enumerate}

We will discuss each of these briefly in turn, in a comparative setting. We
will also look at statistical changepoint analysis, and a divisive
segmentation tree approach.

Unlike in \cite{singhal} we are concerned with segmentation of a
multivariate time series, and not clustering a collection of time series.
This distinction is referred to as, resectively, ``whole clustering'' and
``subsequence clustering'' by \cite{keogh}. Singhal et al.\ \cite{singhal}
build their clustering around principal components analysis, and we also use
Correspondence Analysis -- appropriate for our frequency count data -- to
provide an embedding from which distances can be easily defined.

The great deal of recent work on clustering within time series signals,
e.g.\ the incremental and divisive hierarchical algorithm of 
\cite{rodrigues}, 
has been critiqued by \cite{keogh} on the grounds of meaningfulness.
Goldin et al.\ \cite{goldin} however point out that use of Euclidean
distance is overly simplistic and that data normalization is needed, which
they phrase in terms of ``distances between cluster shapes rather than
between cluster sets''. The need to consider normalization issues very
attentively, and the close resulting links between distance, shape, and
other properties (size, scale, angle, etc.), are all well known in the
Correspondence Analysis area. See \cite{murtaghca}, in particular chapter 4,
which presents case studies of analysis of size and shape, and financial
time series modeling and forecasting.

In this article we develop a new approach to hierarchical clustering of
multivariate time series based on a rigorous treatment of weighting of time
series and attributes.

\subsection{Short Review of Clusterwise Segmentation}

Our innovative approach incorporates piecewise or clusterwise approximation
to a time series or one-dimensional signal. Unlike \cite{spaeth}, we do not
seek to fit patterns directly on the data. Similar type approaches have been
pursued in \cite{murraf,phiros,kamgar,spaeth0}. Instead we fit the piecewise
approximation based on a given hierarchical clustering. Furthermore we do
not want to determine time series segments in a void, removed from external
explanatory and indeed causative discussion. Therefore we are led to regress
the hierarchical segmentation, as an explanatory input, onto time series,
comprising response variables,  that are external to the initial phase of
hierarchical segmentation building.

IsoFinder \cite{isofinder} seeks to segment a genome sequence by repeatedly
finding a splitting point and looking for a significant difference in the
subsequences to the left and right of the splitting point. Significant
difference comes from a t-test between means to left and right; and a
bootstrap-based setting of significance level. One aspect of our approach
that differs from IsoFinder is that we support a fully multivariate input
signal. IsoFinder could remain as a divisive approach for multivariate input
through use of a quadtree and octtree decomposition, although it would
become clumsy relative to our bottom-up tree-building.

Statistical changepoint analysis is exemplified by \cite{fearnhead07}. A
probability model is established for the changepoints. This probability mass
function uses the distance between successive changepoints. Within segments
linear or polynomial regression is used. This is of unknown order which is
to be estimated. Since movement from one changepoint to the next uses
transition probabilities, this is to say that a Markov process is used.
Positions of changepoints and the regression model orders are sought, using
stochastic optimization within a MAP, maximum a posteriori, schema. In \cite%
{fearnhead05}, piecewise linear regression is used for ``regression models
where the underlying functional relationship between the response and the
explanatory variable is modeled as independent linear regressions on
disjoint segments.''

In \cite{gustafsson} applied to speech and motion tracking, even if called
``piecewise constant'' it is in our terminology ``piecewise linear'': the
model is local linear regression with ``piecewise constant parameters''.

In \cite{lemire}, the limits of a clusterwise linear approach are noted. A
``piecewise linear model can determine where the data goes up or down and at
what rate. Unfortunately, when the data does not follow a linear model, the
computation of the local slope creates overfitting.'' In \cite{lemire},
adaptive polynomial (including a 0 order, or constant) degree of fit in each
piecewise region is pursued in the modeling of stock prices and ECG data.

Relative to \cite{fearnhead07} and, for example, \cite{stephens}, our
approach is not parametric. In terms of evaluation therefore we are more
concerned to link results with external data, both qualitative (e.g.\ causes
of changepoints of varying degrees of importance) and quantitative (e.g.\
other univariate time series). Therefore the applications of our work and
their context is somewhat different from these parametric modeling
approaches.

\subsection{Structure of this Article}

In this article, we will interchangeably refer to a time series defined on a
timeline and a signal with support given by this timeline. Broadly speaking
we will also use interchangeably the following terms: hierarchy, tree,
dendrogram, ultrametric space. A range of other terms, algorithms and
methods are introduced and discussed in the Appendices.

In section \ref{sect2} the data and problem is discussed, together with
aspects relating to methodology that are at issue in this article.

In section \ref{sect3}, the relevance and indeed importance of hierarchy in
the data analysis is addressed.

In section \ref{sect4} the association of a hierarchy with an external
signal is at issue. We develop a new way to map the latter onto the
hierarchy.

In section \ref{sect5} we study how the mapping of external signal to the
hierarchy can lead to new insights.

\section{Background of the Data}

\label{sect2}

The data used in this work related to conflict violence and were produced by
CERAC, the Conflict Analysis Resource Center, www.cerac.org.co. We will
refer to the data used as the CERAC Colombia Conflict database. It is being
grown on an ongoing basis. From \cite{rsv1}, we use high frequency
micro-data, relating to internal conflict violence in Colombia (population:
44 million) from 1988 to 2004, hence over a period of more than 16 years.

Conflict violence in Colombia is not based primarily on ethnic, religious,
or regional differences, as is often the case elsewhere, but it instead has
roots in economic and political factors. Economic drivers include, for
instance, the narcotics sector and kidnapping. 
The CERAC data records discrete action types, their intensity, dates,
locations and other information. Event types are broken down at the first
divide into clashes requiring multilateral engagement, and attacks which
are unilateral. Intensity breaks down mainly into killings and injuries. For
the data period, the Colombian conflict has seen more than 3000 killings
plus injuries per year. 
One use of the dataset has been to analyze civilian casualties \cite{rsv3}.

In \cite{johnson} distribution of fatalities over time was found to follow a
power law. Furthermore the work of these authors points to how remarkably
similar power law behavior can be found in conflicts ranging from Iraq to
Indonesia. Such power law behavior leads to self-similarity, i.e.,
burstiness on all aggregation scales. (See examples of application of this
in \cite{aussem} and general discussion in \cite{newman}.) As a result it is
difficult to smooth such data, and hence it is difficult to fit simple
parametric models. Under such circumstances it is feasible to seek and find
patterns and trends in the data, and it is precisely these goals which are
at issue in this work.

In this article, we present a novel approach for the study of dynamics.
Firstly we look at visualizations using Correspondence Analysis \cite%
{murtaghca}. The output spatial representation is an equiweighted Euclidean
one, with embedding of both rows and columns -- events, and the attributes
used for these events -- and this allows cluster analysis and other analyses
to bypass any other form of data normalization or standardization.

We then study change versus continuity over time, allowing for gradation in
such change. A hierarchical clustering is used, taking account of the
timeline, presenting change cluster, breakpoints and timeline
resolution-related properties.

Context is offered by the following. In \cite{crane}, the price of cocaine
in the US from 1983 onwards 
is found to be affected by interdiction effectiveness,
and this in turn is found to be beneficial in terms of reduced consumption,
at least in the short run.





In our analysis we will look at how narcotics prices might be related to the
Colombian violence. Causal mechanisms could take place in different ways.
For example a declining world market for narcotics could put great pressure
on production, leading to violent reaction when faced with declining
returns. Or an expanding market could lead to greater violence because the
spoils are greater. A further consideration is that causal mechanisms may
work themselves out at different rates, and even in different ways, on the
wholesale and retail markets. The retail markets for illegal narcotics are
driven by actors who are distributed far and wide relative to the locations
of production.

The cocaine drug comes from the \emph{coca} plant. Most initially processed
cocaine on the world narcotics markets come from Colombia. A main tool in
combating production by the Colombian government has been crop spraying,
leading to production being dispersed in jungle regions or national parks.
The coca plant is very different from the \emph{cocoa} bean from the \emph{%
cacao} tree, from which chocolate is produced. \emph{Coffee}, from coffee
beans of the coffee plant, is the leading legal agricultural export from
Colombia. The latter fact is important because (for example, as just one
possible economic mechanism) a declining market for coffee could conceivably
lead to product substitution in response.

Opiates come from the opium poppy. Like cocaine, opiates are also narcotic
alkaloids. Morphine, derived from opium, is used for medicinal purposes,
whereas heroin is a generally banned narcotic. While poppy production has
been at times of some significance in Colombia, mostly the leading opiates
producer worldwide has been Afghanistan. However for supply of opiates 
to the US, Colombia and Mexico play a major role.  
As noted above in regard to the
complexity of economic mechanisms, there are intricate linkages between,
say, cocaine and opiates, involving many social actors.

\section{Hierarchy: Tracking and Prioritizing of Change}

\label{sect3}

\subsection{Data}

\label{data}

We used 144 numerical attributes relating to killings and injuries coming
from 20,288 dated events: see Appendix 1.  Aggregating by month the
numerical data, 144 attributes used, yielded data for 204 successive months.
Further aggregating by year provided us with data for the 15 years, 1990 to
2004, each year having a 144-valued vector of values where we eliminated
1988 and 1989 because we did not have narcotics data for these years (see
section \ref{sect4}).

A short periodization of the conflict is as follows \cite{rsv1}:
reintensification of conflict in 1986, roughly constant then to 1994,
followed by continuous acceleration. 1988--1991, \textquotedblleft
adjustment period\textquotedblright\ due to the end of the Cold War;
1992--1996, \textquotedblleft stagnation period\textquotedblright ; and from
1997 onwards, \textquotedblleft upsurge period\textquotedblright . In the
section to follow there will be further discussion of periodization.

\subsection{Considerations on the Analysis}

\label{considerations}

\begin{figure}[tbp]
\begin{center}
\includegraphics[width=12cm]{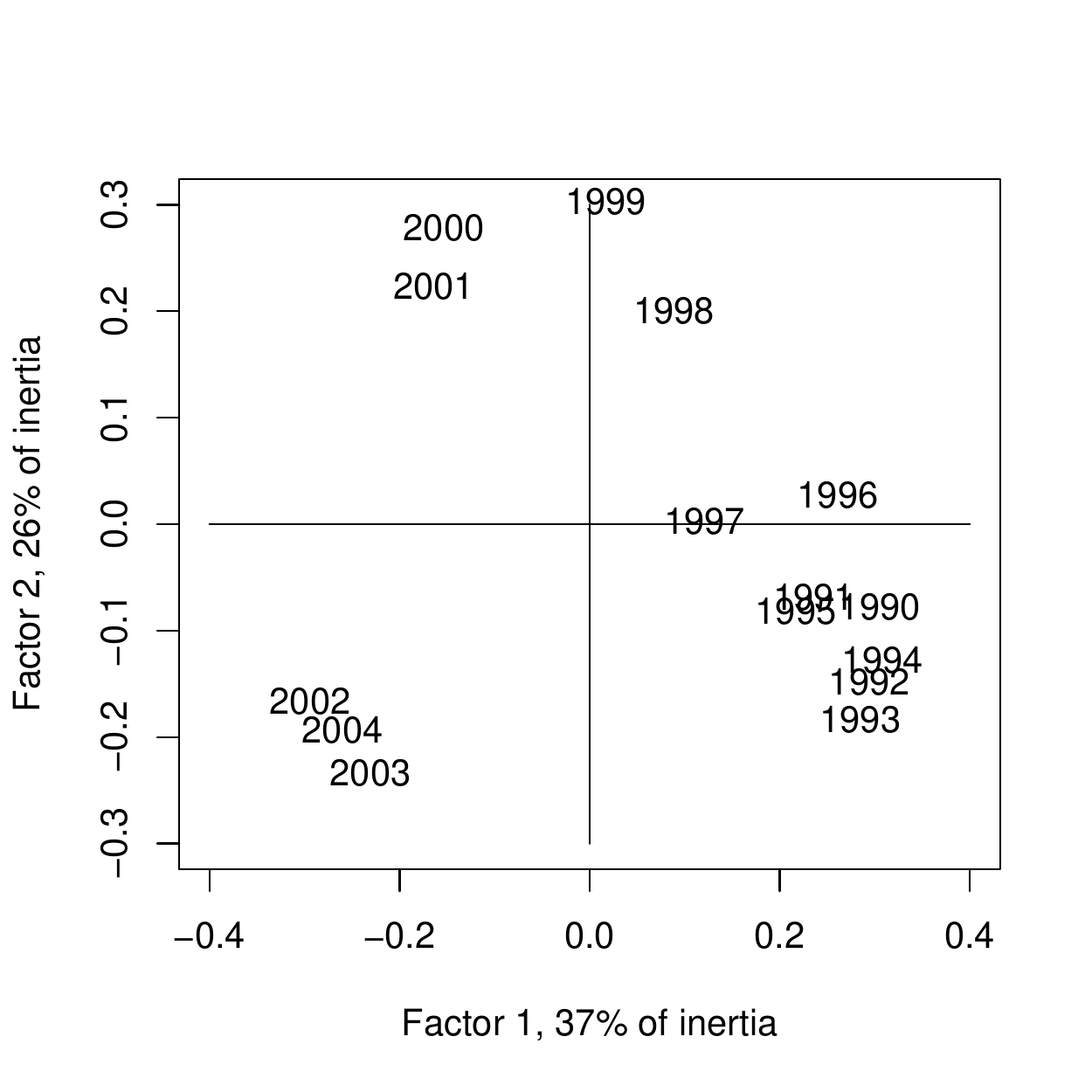}
\end{center}
\caption{Correspondence Analysis of the annual aggregated data using 144
attributes. Years shown. As discussed in the text, gaps representing the
more important changepoints are (i) between 1997 and 1998; and (ii) between
2001 and 2002. }
\label{fig1}
\end{figure}

Figure \ref{fig1} shows a best planar representation, accounting for 62\% of
the information content of the data as expressed by the combined moments of
inertia of the cloud of years (or of the cloud of attributes) about these
two axes. Following somewhat clumped characteristics of the years 1990
through 1997, there is a break then with the year 1998. From 1997 an arc is
formed, up to 2004. A break between the years 1997 and 1998, (i), is
followed by a later break, between 2001 and 2002, (ii).

The 1998 changepoint, (i), was in a period when guerrilla gains from 1996
onwards were being reversed, a process that was seen to be so by 1999. There
were high levels of government casualties in 1997--1998, due to guerrilla
operations against isolated military and police bases. In particular through
airborne weaponry, the government got the upper hand. The ``upsurge period''
from 1997 onwards also saw a rise of (anti-guerrilla) paramilitary activity
whereas before they had been involved in drug trafficking. The
paramilitaries started operations around 1997. There was a consolidation of
paramilitary groups in 1997, announced publicly in December 1997, and they
became active then, having many of their own number killed. It was not until
1999 that paramilitaries began to kill large numbers of guerrillas. Among
all of these mutually influencing, reinforcing or retarding, trends and
events, our analysis points to a succession of two years where the global
change was most intense.

We now come to the 2002 changepoint, (ii). A peak of (paramilitary)
casualties brought about by government against paramilitaries was in 2002
due to aerial bombardment of a paramilitary position under attack by
guerrillas. It was a major setback for the paramilitaries, who declared a
truce following the election that year of President Alvaro Uribe.

\begin{figure}[tbp]
\begin{center}
\includegraphics[width=8cm]{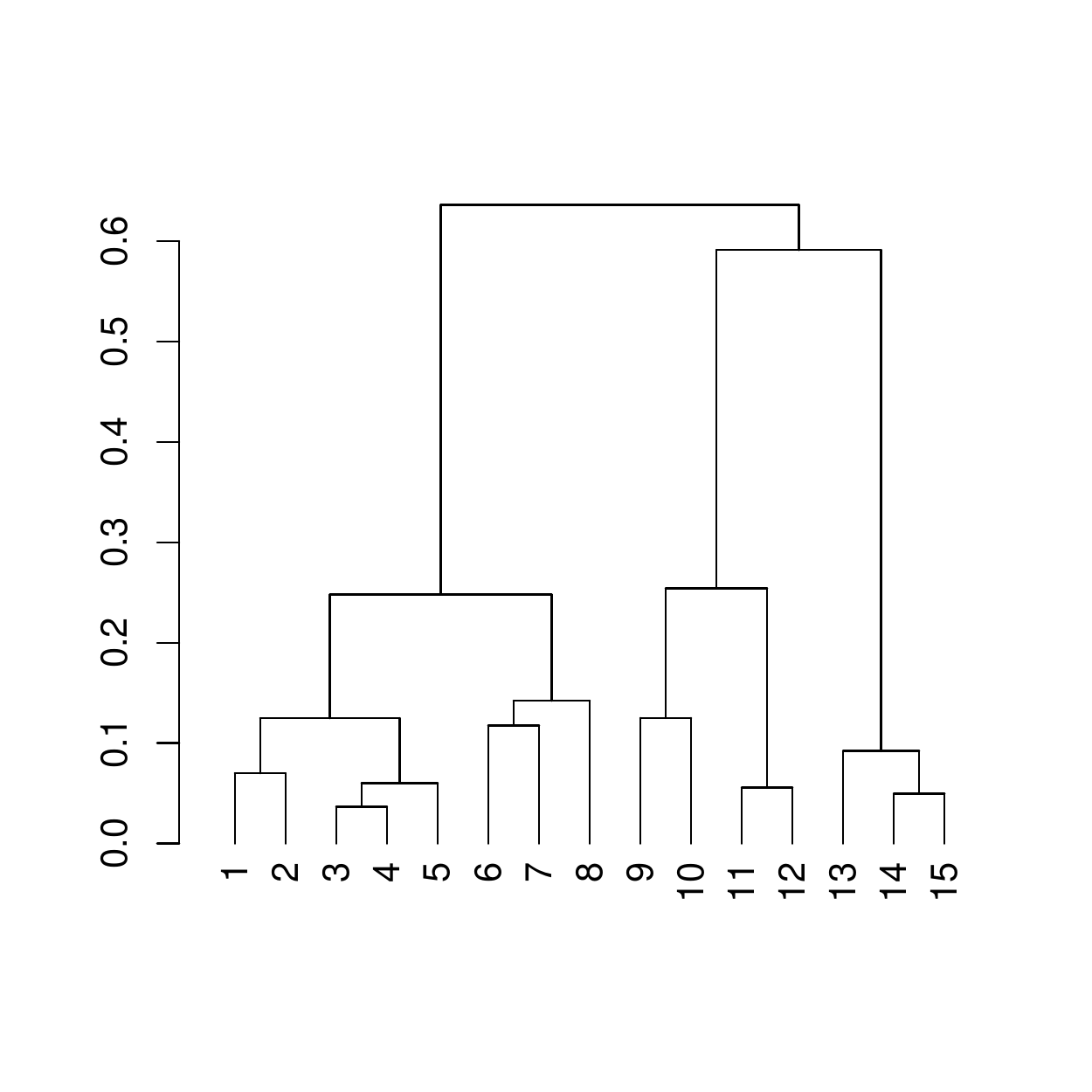}
\end{center}
\caption{Hierarchical clustering of the fifteen years from 1990 to 2004,
using the principal factor values from Figure \protect\ref{fig1}. A sequence
constrained complete link agglomeration criterion was used. The main
caesuras or changepoints here are (i) in moving from year 8 = 1997 to year 9
= 1998; and (ii) in moving from year 12 = 2001 to year 13 = 2002.}
\label{fig2}
\end{figure}

Figure \ref{fig2} shows (i) the caesuras between 1997 (8th year) and 1998
(9th year); and (ii) between 2001 (12th year) and 2002 (13th year). Subtrees
before and after these caesuras are clearly distinguishable in the full
tree. In line with Figure \ref{fig1}, what Figure \ref{fig2} indicates is
that these are the important caesuras in this data. Note that the same data
as seen in the planar projection in Figure \ref{fig1}, viz.\ the factor 1
and 2 projections, is used for the construction of the hierarchy of Figure %
\ref{fig2}.

The data used to construct the hierarchy in Figure \ref{fig2} is of inherent
dimension min(20288 events, 204 months, 144 attributes, 15 years) $- 1$
(from section \ref{data}; see also Appendices 1 and 2). Furthermore, based
on Figure \ref{fig1}, we in fact used the first two factors only. So the
input data dimensionality on which the hierarchy in Figure \ref{fig2} is
based is 2. 

\begin{figure}[tbp]
\begin{center}
\includegraphics[width=8cm]{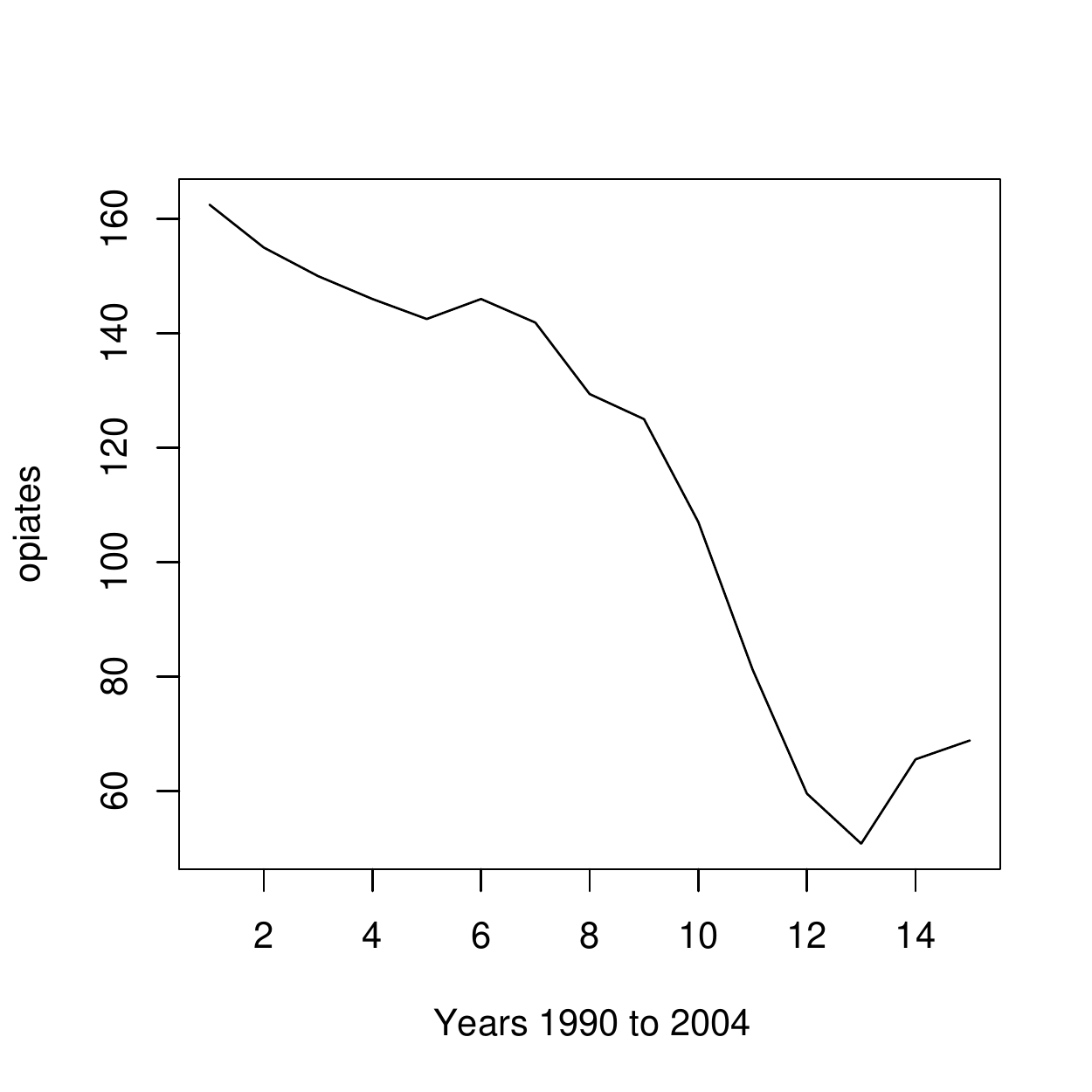}
\end{center}
\caption{Wholesale street prices of opiates in the US over 15 years, 1990 to
2004. (Source of data: \protect\cite{opiates}.)  Ordinate units are US\$/kg.}
\label{fig3}
\end{figure}

\subsection{Novelties in Methodology}

Correspondence Analysis (Figure \ref{fig1}) embeds both annually aggregated
events and their attributes in the same Euclidean space. 
As a basis for subsequent analysis, e.g.\ clustering, weighting and
normalization are taken care of as part of the Euclidean embedding
algorithm. Appendix 2 presents an overview of Correspondence Analysis.

The hierarchical clustering of Figure \ref{fig2} is built on the timeline.
This means that clusters are contiguous on the yearly timeline. Input to the
hierarchical clustering is factor projections from the Correspondence
Analysis implying that the 15 yearly aggregated events are equiweighted and
endowed with the Euclidean distance. Appendix 3 presents an overview of this
sequence-constrained hierarchical clustering.

Figure \ref{fig2} provides a hierarchical understanding of the Colombian
conflict violence over the years 1990 to 2004. It ranks change by order of
importance. Bigger change is associated with more distinct branches in the
tree. A hierarchy defines an ultrametric topology. So we can justifiably
characterize Figure \ref{fig2} as an ultrametric, or tree metric,
understanding of the Colombian conflict violence. In the same way the
semantics of Colombian society are displayed in a metric, rather than
ultrametric, way in Figure \ref{fig1}.

\section{Association between Hierarchical Understanding and External Signal}

\label{sect4}

\subsection{Regressing Opiates Wholesale Prices Response Variable on
Explanatory Hierarchy}

We have a structured view in Figure \ref{fig2} of the socio-dynamics of the
situation in Colombia over the years 1990 to 2004. Let us call it $\mathcal{H%
}$. This is the set of interrelationships expressed in Figure \ref{fig2}
over the years 1990--2004. It is a summarized view of those years. It is a
synthetic view, and depends not just on the algorithms applied but -- a most
important aspect -- on the input data used. As a structured view of the
socio-dynamics of Colombia, we now seek to associate $\mathcal{H}$ with
other external signals defined on the same time period.

As potentially relevant economic sectors outside Colombia we will consider
the narcotics sector. We will consider two narcotics signals.

Figure \ref{fig3} shows the opiates wholesale price, in the US, averaged per
year, in US\$/kg. Our aim is to look for linkages between the evolving
Colombian socio-dynamic expressed in Figure \ref{fig2} and distant
reflections of this, such as the particular narcotics market situation shown
in Figure \ref{fig3}. Production is expressed (in a highly complicated way)
in the former, Figure \ref%
{fig2}, while consumption is expressed in the latter, Figure \ref{fig3}.

As noted in section \ref{considerations} there are particular breakpoints
easily visible in Figure \ref{fig2}, e.g.\ in moving from year 8 = 1997 to
year 9 = 1998; and in moving from year 12 = 2001 to year 13 = 2002. In
Figure \ref{fig3} something of this is visible too: year 9 is early in a
period of precipitous fall (i.e., fall in wholesale price); and year 13 is a
clearly visible turning point.

Let us map the opiates data in Figure \ref{fig3} into the hierarchical
structure, $\mathcal{H}$, of Figure \ref{fig2}. One way to do this is to
have a set of within-cluster constant approximations to the opiates (Figure %
\ref{fig3} data), and furthermore to allow for a range of such constant
approximations.

In practice we will use constant approximations of the opiates data, Figure %
\ref{fig3}, that are governed by the embedded clusters of $\mathcal{H}$,
Figure \ref{fig2}. However we will not proceed by cluster level (or
dendrogram level from the root) because this would imply different cluster
cardinalities at each level -- and hence a not very sensible ordering of
what are ultimately approximations to the full data set. We will read off
approximations to the hierarchy by decreasing \emph{differences} in
information between dendrogram levels. We will allow ourselves to be guided
by these differences in information between levels. This information is
available to us through the dendrogram's Haar wavelet transform. See
Appendix 4 for background detail.

In Figure \ref{fig5}, we see the succession of better approximations,
quantified by mean squared error (MSE), as given by decreasing absolute
values of the dendrogram Haar wavelet detail coefficients.

In regard to use here and later of MSE, note that the size of this goodness
of fit value is strongly influenced by the signal's values, -- cf.\
ordinates in displays. The MSE is not scale-invariant and the ordinate scale
impacts on the MSEs.

\begin{figure}[tbp]
\begin{center}
\includegraphics[width=12cm]{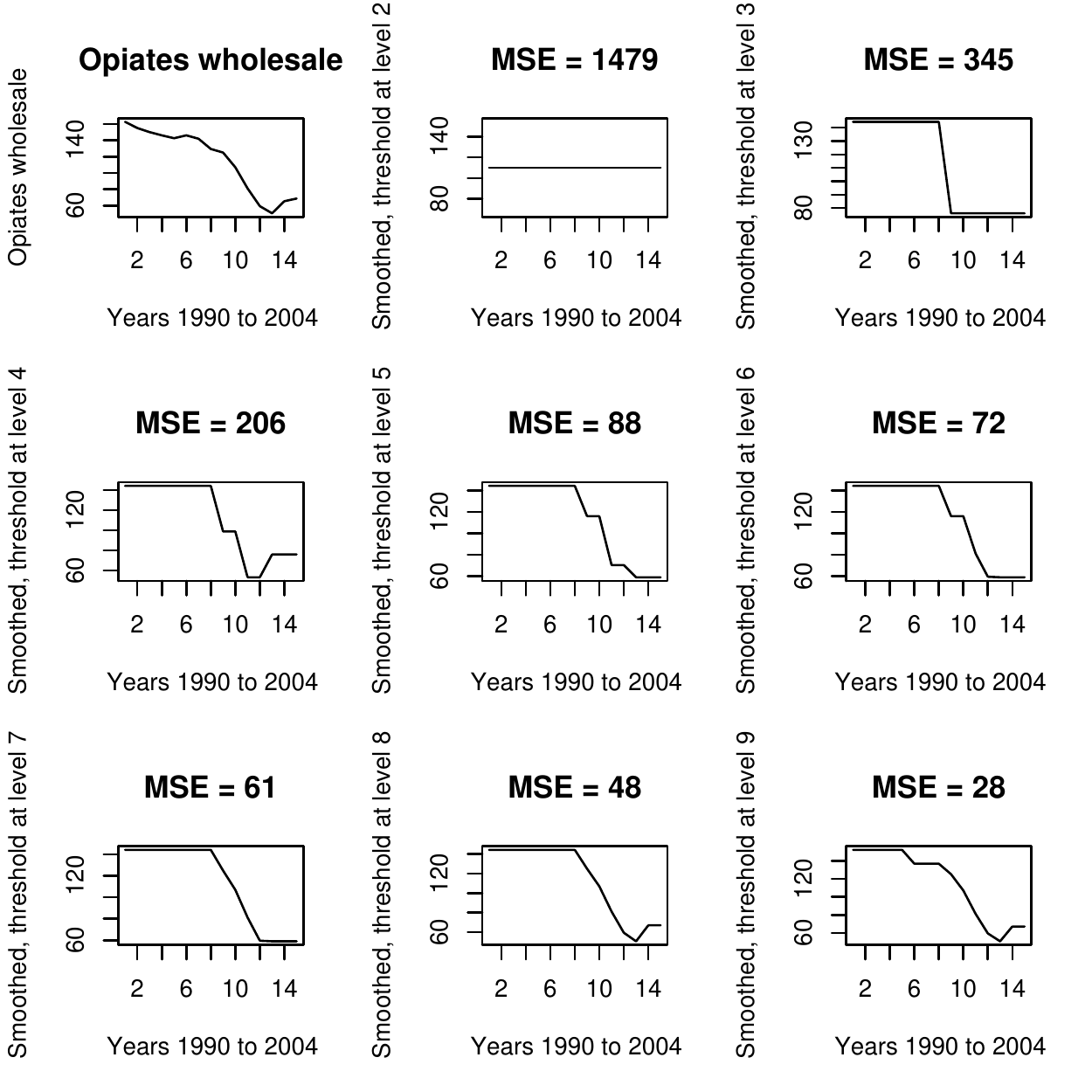}
\end{center}
\caption{Upper left: opiates, wholesale prices, over the 15 years, 1990 to
2004. Then from left to right, row-wise: reconstruction of the data based on
thresholding using decreasing absolute values of the dendrogram wavelet
detail coefficients. The MSE, mean square error, indicates the quality of
approximation (with a small value being best).}
\label{fig5}
\end{figure}

Note that the indication ``thresholded at level 3'', for instance, in Figure %
\ref{fig5}, is to be understood as at the third, ordered by decreasing
value, wavelet detail coefficient. This is not necessarily the third
dendrogram level from the root.

\subsection{Further Response Variables}

\begin{figure}[tbp]
\begin{center}
\includegraphics[width=12cm]{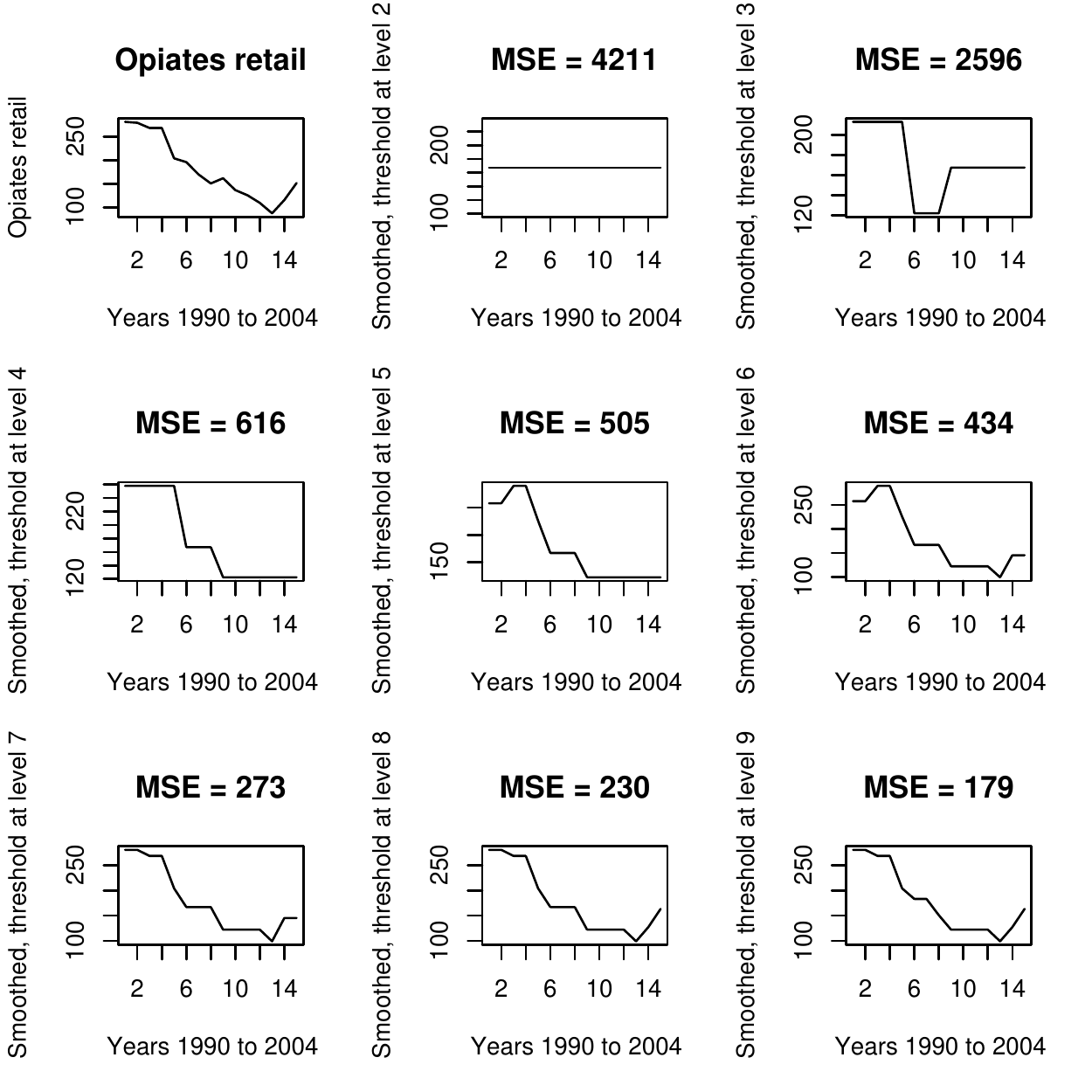}
\end{center}
\caption{Upper left: opiates, retail prices, over the 15 years, 1990 to
2004. Then from left to right, row-wise: reconstruction of the data based on
thresholding using decreasing absolute values of the dendrogram wavelet
detail coefficients. The MSE, mean square error, indicates the quality of
approximation.}
\label{fig6}
\end{figure}

\begin{figure}[tbp]
\begin{center}
\includegraphics[width=12cm]{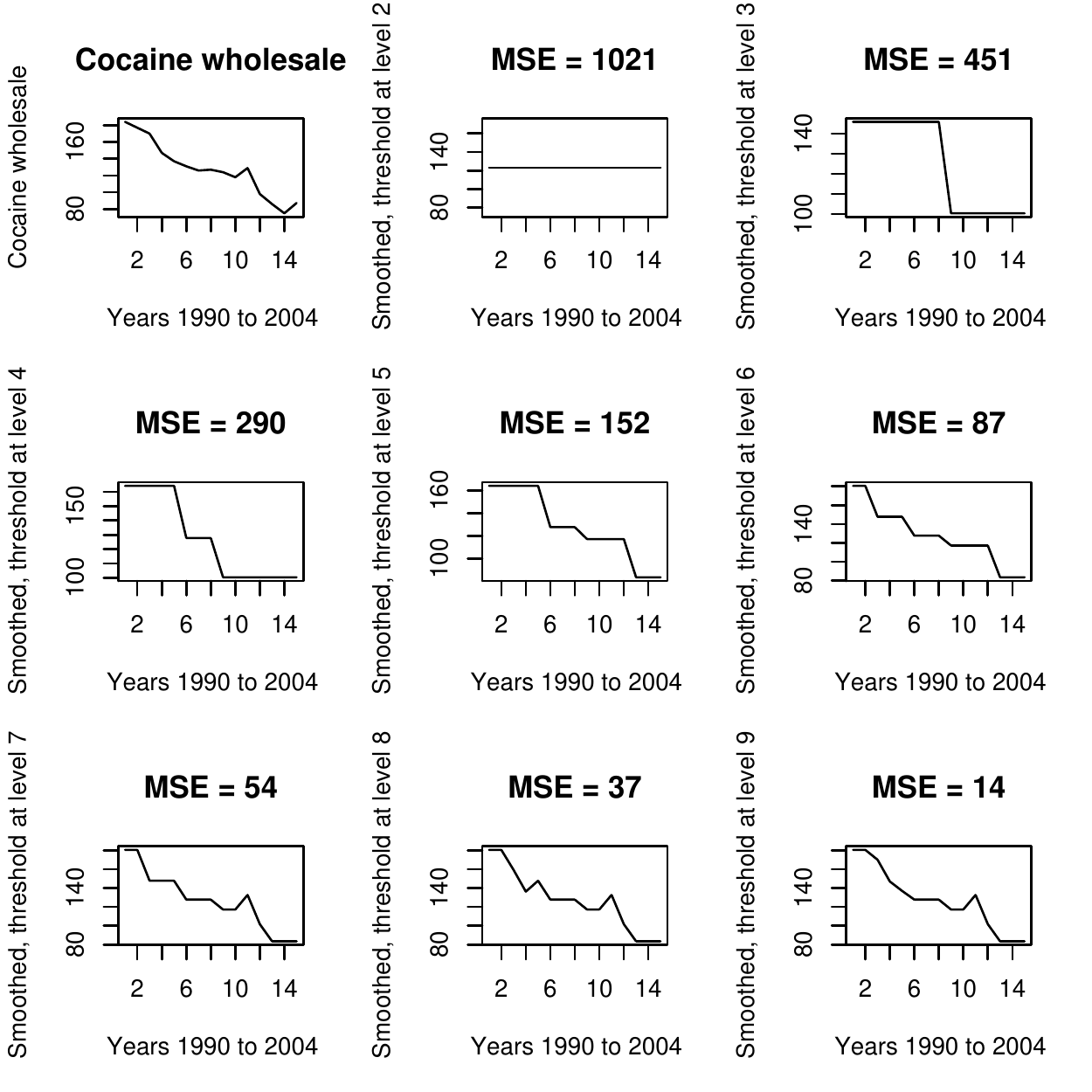}
\end{center}
\caption{Upper left: cocaine, wholesale prices, over the 15 years, 1990 to
2004. Then from left to right, row-wise: reconstruction of the data based on
thresholding using decreasing absolute values of the dendrogram wavelet
detail coefficients. The MSE, mean square error, indicates the quality of
approximation.}
\label{fig7}
\end{figure}

\begin{figure}[tbp]
\begin{center}
\includegraphics[width=12cm]{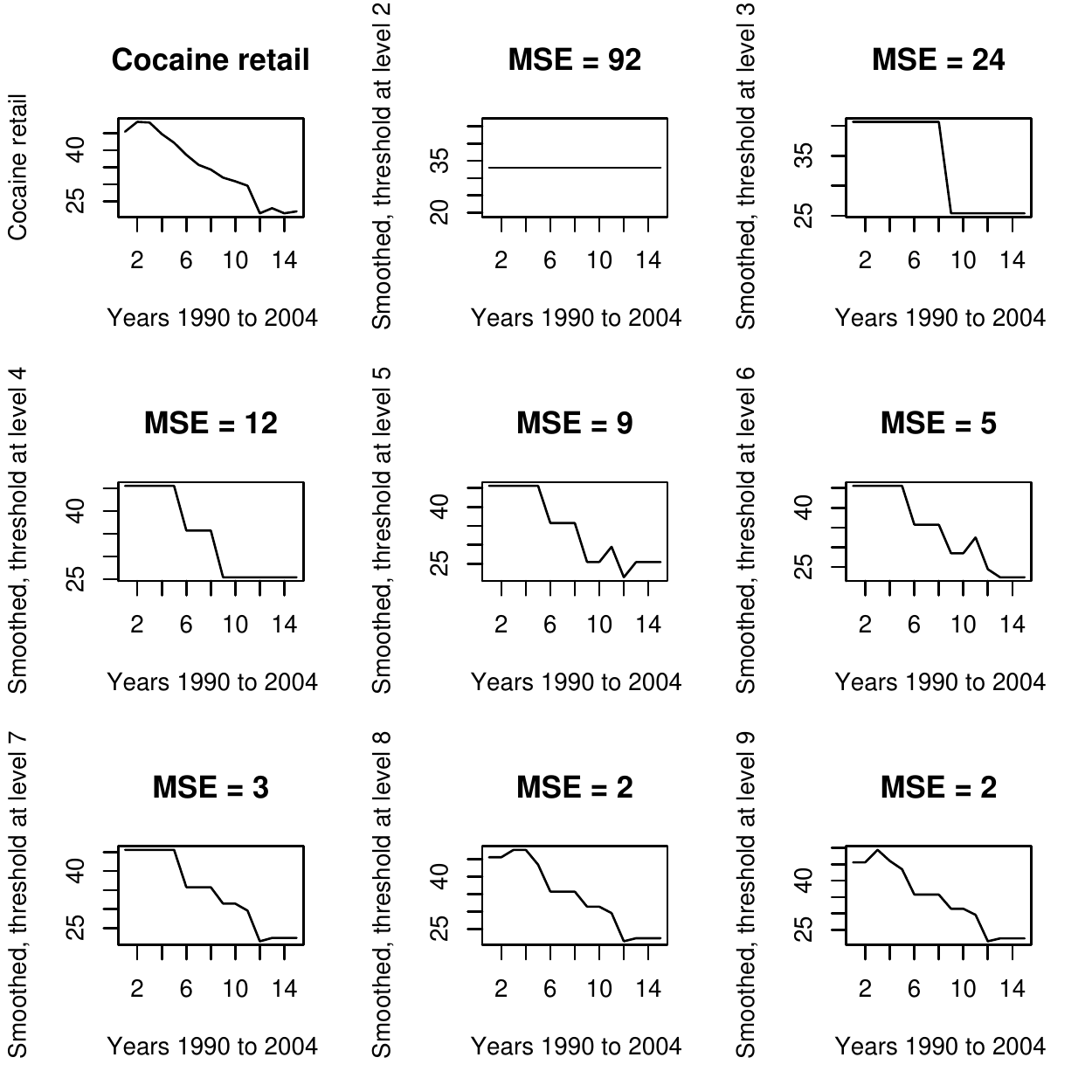}
\end{center}
\caption{Upper left: cocaine, retail prices, over the 15 years, 1990 to
2004. Then from left to right, row-wise: reconstruction of the data based on
thresholding using decreasing absolute values of the dendrogram wavelet
detail coefficients. The MSE, mean square error, indicates the quality of
approximation.}
\label{fig8}
\end{figure}

We additionally now consider a set of signals, see upper left panel in
Figures \ref{fig5}, \ref{fig6}, \ref{fig7} and \ref{fig8}. There are,
including the initial analysis of the previous section:

\begin{itemize}
\item Opiates and cocaine retail prices (street price), US\$/gram, see \cite%
{opiates}.

\item Opiates and cocaine wholesale price, US\$/kg, see \cite{opiates}.
\end{itemize}

In Figures \ref{fig5}, \ref{fig6}, \ref{fig7} and \ref{fig8} exact
reconstruction of the input signal, displayed in the upper left, is possible
if no thresholding is used.

Figure \ref{fig6} shows the successive approximations to the opiate retail
prices -- again in the US, from 1990 to 2004, street prices in US\$/gram 
\cite{opiates}.

Figure \ref{fig7} shows the successive approximations to the cocaine
wholesale prices -- in the US, from 1990 to 2004, in US\$/kg \cite{opiates}.

Figure \ref{fig8} shows the successive approximations to the cocaine retail
prices -- in the US, from 1990 to 2004, street prices in US\$/gram \cite%
{opiates}.

\subsection{Novelties in Methodology}

We denote the hierarchy in Figure \ref{fig2} as $\mathcal{H}$ and a signal
such as that in Figure \ref{fig3} as $\mathcal{s}$. Both $\mathcal{H}$ and $%
\mathcal{s}$ have the same support, viz.\ the ordered years 1990 to 2004. We
are projecting the ultrametric space, $\mathcal{H}$, onto the signal $%
\mathcal{s}$.  We have: $f: \mathcal{H \longrightarrow s}$. This can be
viewed too as an ultrametric regression, i.e.\ we are regressing $\mathcal{s}
$ on $\mathcal{H}$. We estimate our signal from the hierarchy: $\widehat{ 
\mathcal{s }} = f(\mathcal{H)}$.

Consider now the segmentation as a clusterwise regression problem. We could
approach the clusterwise regression by partition that is read off the
hierarchy in Figure \ref{fig2}. Using the Haar wavelet transform instead
reads off partitions in order of MSE (mean square error) of approximation of
the input signal by the wavelet filtered and reconstructed signal. The MSE
provides us with a measure of the approximation of the piecewise constant
(which is also linear: but for constant ordinate or response variable) fit
to the input signal. Note that each partition of the hierarchy covers the
full signal, i.e.\ each partition is by definition defined on the input's
support.

\section{Patterns in Data Reflecting the Known Hierarchical Structuring of
Events}

\label{sect5}

\subsection{Smoothing based on Ultrametric Wavelet Transform}

\begin{figure}[tbp]
\begin{center}
\includegraphics[width=12cm]{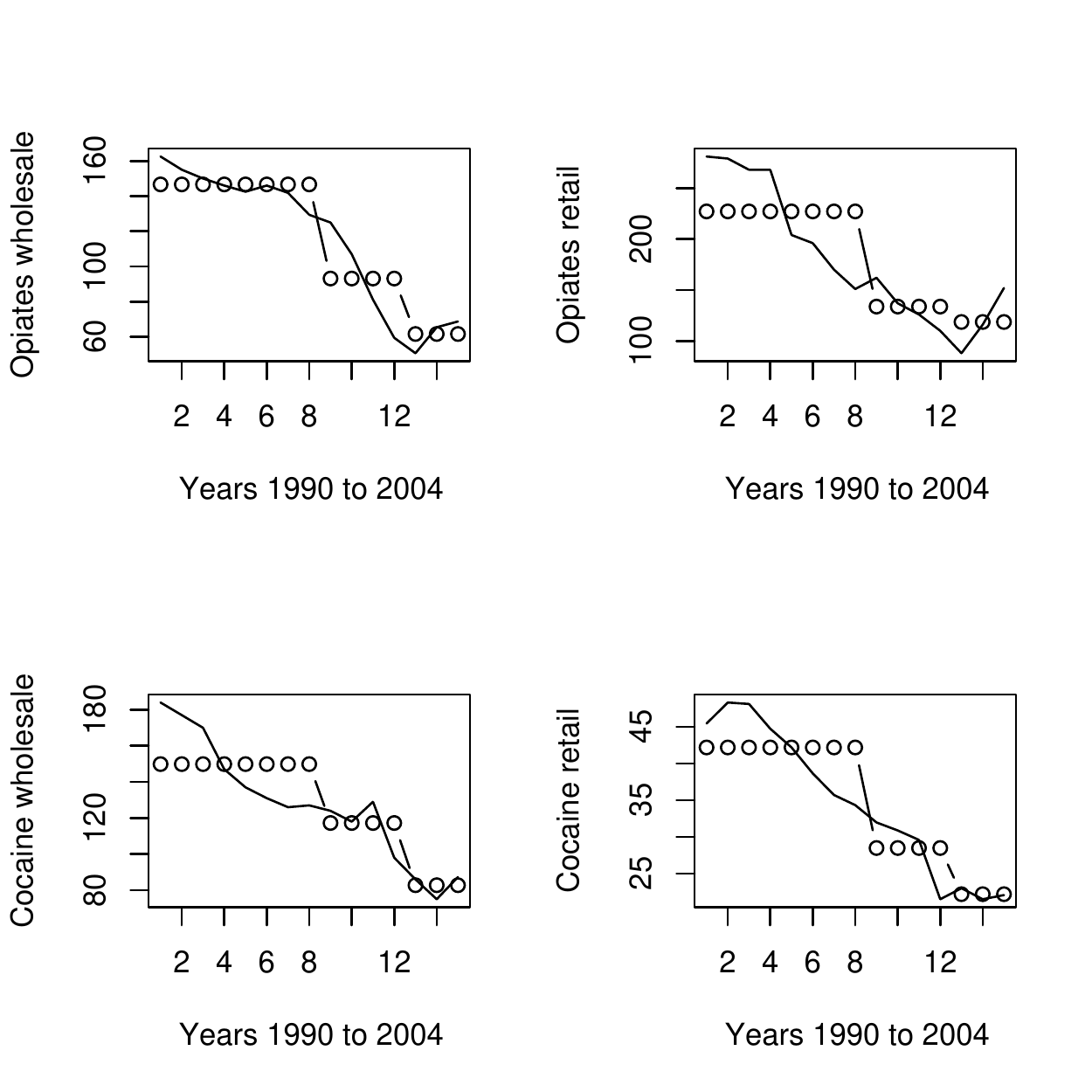}
\end{center}
\caption{We take the three clearest clusters in Figure \protect\ref{fig2},
viz.\ from year 1 = 1990 to year 8 = 1997; from year 9 = 1998 to year 12 =
2001; and from year 13 = 2002 to year 15 = 2004. Then we simply trace the
piecewise constant fits, in each case, to the four different narcotics
market signals. These fits are shown by the small circle piecewise constant
curves. The mean square errors are, respectively for opiates wholesale and
retail, and cocaine wholesale and retail: 223, 1533, 305 and 18.}
\label{fig8888}
\end{figure}

A baseline scenario for what we now seek is shown in Figure \ref{fig8888}.
In this figure we show a simple and direct clusterwise fit to each of the
four signals. We replace the first eight values, in the first cluster, with
their mean value; and similarly for the second cluster; and the third. The
goodness of fit is given for each quadrant from upper left to lower right as
indicated in the figure caption.

Can our wavelet regression perform better? The mean square errors (MSEs), we
find, are improved in all four cases. Respectively, as we will see, the MSE
of opiates wholesale decreases from 223 to 206. The MSE of opiates retail
decreases from 1533 to 616. The MSE of cocaine wholesale decreases from 305
to 290. And the MSE of cocaine retail decreases from 18 to 12.

\begin{figure}[tbp]
\begin{center}
\includegraphics[width=12cm]{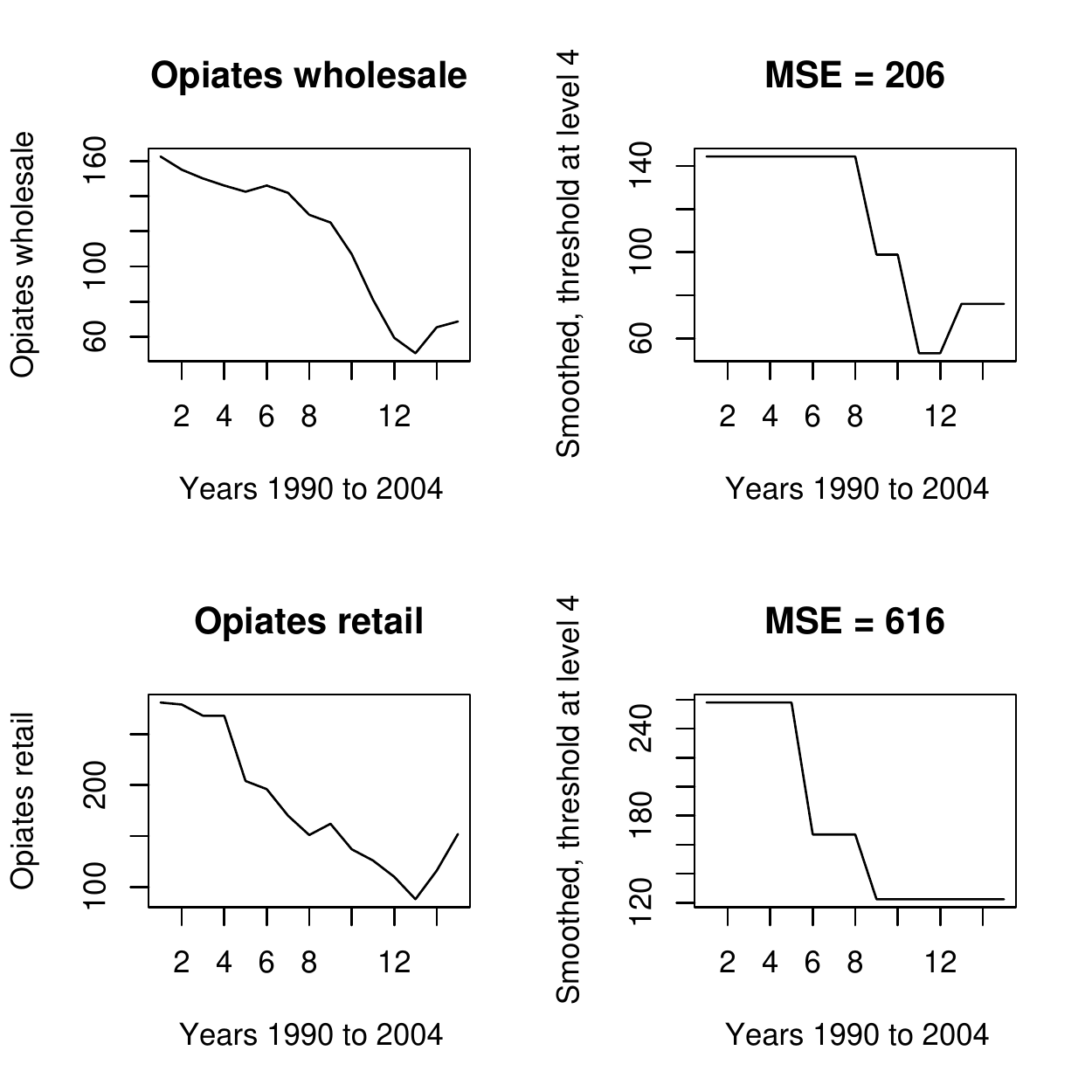}
\end{center}
\caption{Left: opiates, wholesale and retail prices, over the 15 years, 1990
to 2004. Right: corresponding reconstruction of the signal based on wavelet
detail thresholding. (The two upper panels are identical to panels shown in
Figure \protect\ref{fig5}; and the two lower panels are identical to panels
shown in Figure \protect\ref{fig6}.)}
\label{fig9}
\end{figure}

\begin{figure}[tbp]
\begin{center}
\includegraphics[width=12cm]{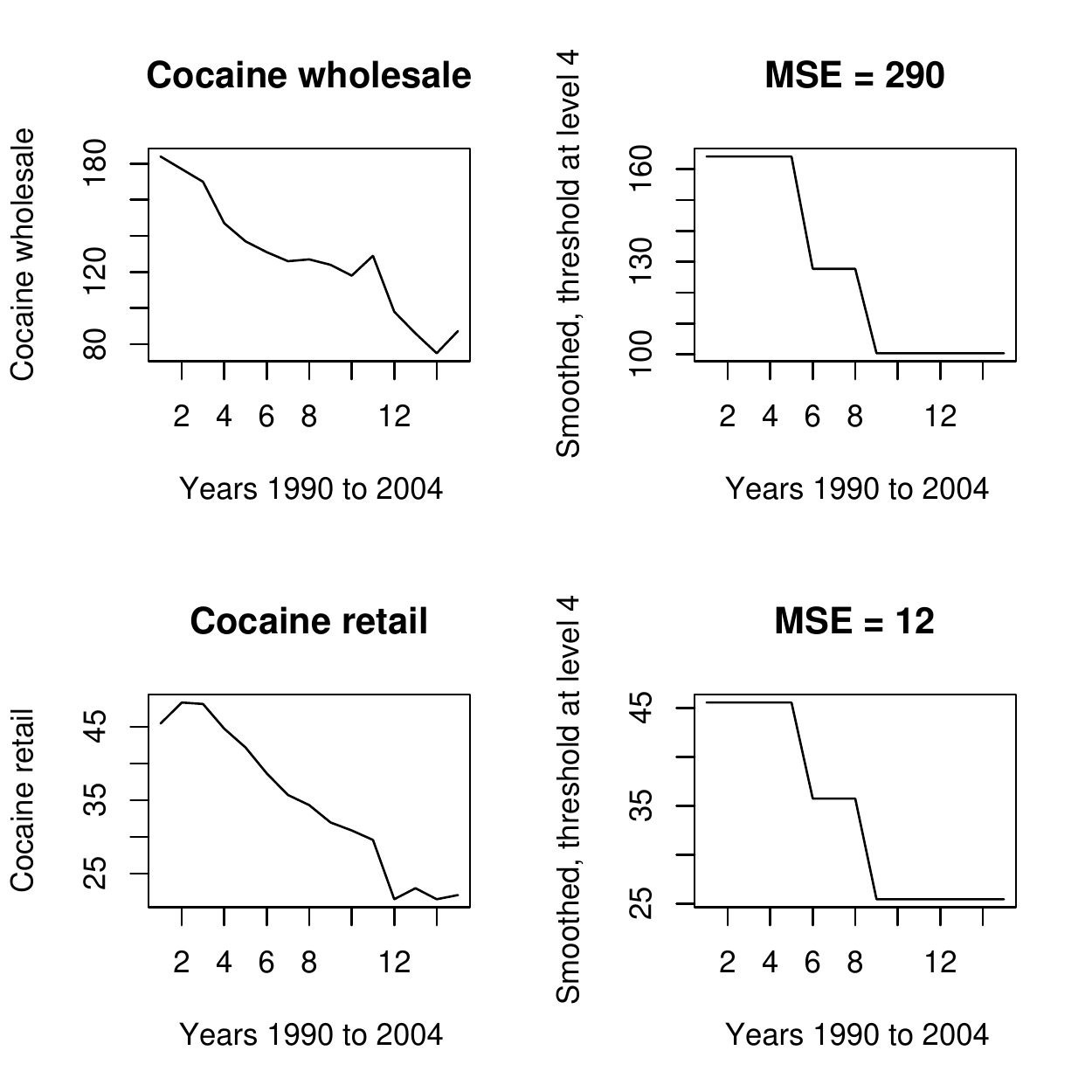}
\end{center}
\caption{Left: cocaine, wholesale and retail prices, over the 15 years, 1990
to 2004. Right: corresponding reconstruction of the signal based on wavelet
detail thresholding. (The two upper panels are identical to panels shown in
Figure \protect\ref{fig7}; and the two lower panels are identical to panels
shown in Figure \protect\ref{fig8}.)}
\label{fig10}
\end{figure}

From Figures \ref{fig5}, \ref{fig6}, \ref{fig7} and \ref{fig8} we selected
segmentations with at least three segments in each, corresponding to the
major breakpoints discussed in section \ref{considerations}. The signals
used and the clusterwise constant approximated versions are shown in Figures %
\ref{fig9} and \ref{fig10}.

Figures \ref{fig9} and \ref{fig10} can be viewed as follows. From the
hierarchy in Figure \ref{fig2}, with its important caesuras as discussed in
section \ref{considerations}, take a range of within-cluster signal values
as their mean. This implies that we have a piecewise linear approximation of
the signals, in the sense of clusterwise constant. Compared to our baseline
scenario shown in Figure \ref{fig8888} we intervene on -- modify in wavelet
transform space -- a number of clusters characterized by small changes in
the hierarchy.

Our baseline of Figure \ref{fig8888} shows breakpoints over the 15-year 
time span.  The respective breakpoints of Figures \ref{fig9} and \ref{fig10}
are largely consistent, as summarized in Table \ref{tabbreakpoints}, 
while (we may say informally) 
using the hierarchy as a ``key'' or ``template''.  

\begin{table}
\begin{center}
\begin{tabular}{|llccrlccc|} \hline
Crude 3 segments & Fig.\ \ref{fig8888} & 1 & & -- 8 & 9 -- 12 & & 13 -- & 15\\
Opiates wholesale & Fig.\ \ref{fig9} & 1 &   & -- 8 & 9 -- 10 & 11 -- 12 & 13 -- & 15 \\
Opiates retail    & Fig.\ \ref{fig9} & 1 & -- 5 & 6 -- 8 & 9 -- &      &    & 15  \\
Cocaine wholesale & Fig.\ \ref{fig10} & 1  & -- 5 & 6 -- 8 & 9 -- &      &      & 15  \\
Cocaine retail    & Fig.\ \ref{fig10} & 1 & -- 5 & 6 -- 8 & 9 -- &      &     & 15  \\ 
\hline
\end{tabular}
\end{center}
\caption{Breakpoints in the Colombian 15-year violence timeline.
The breakpoints were read off the figures noted in the Table.
We see, for example, that for the cocaine wholesale signal, the 
breakpoints are such that there is a segment consisting of years 1 to 5; 
then a segment of years 6 to 8; and finally a segment from year 9 to 
year 15.  
From the crude signal segmentation of Figure \ref{fig8888} and the 
wavelet smoothings of Figures \ref{fig9} and \ref{fig10}, there is
considerable consistency.}
\label{tabbreakpoints}  
\end{table}

\begin{figure}[tbp]
\begin{center}
\includegraphics[width=12cm]{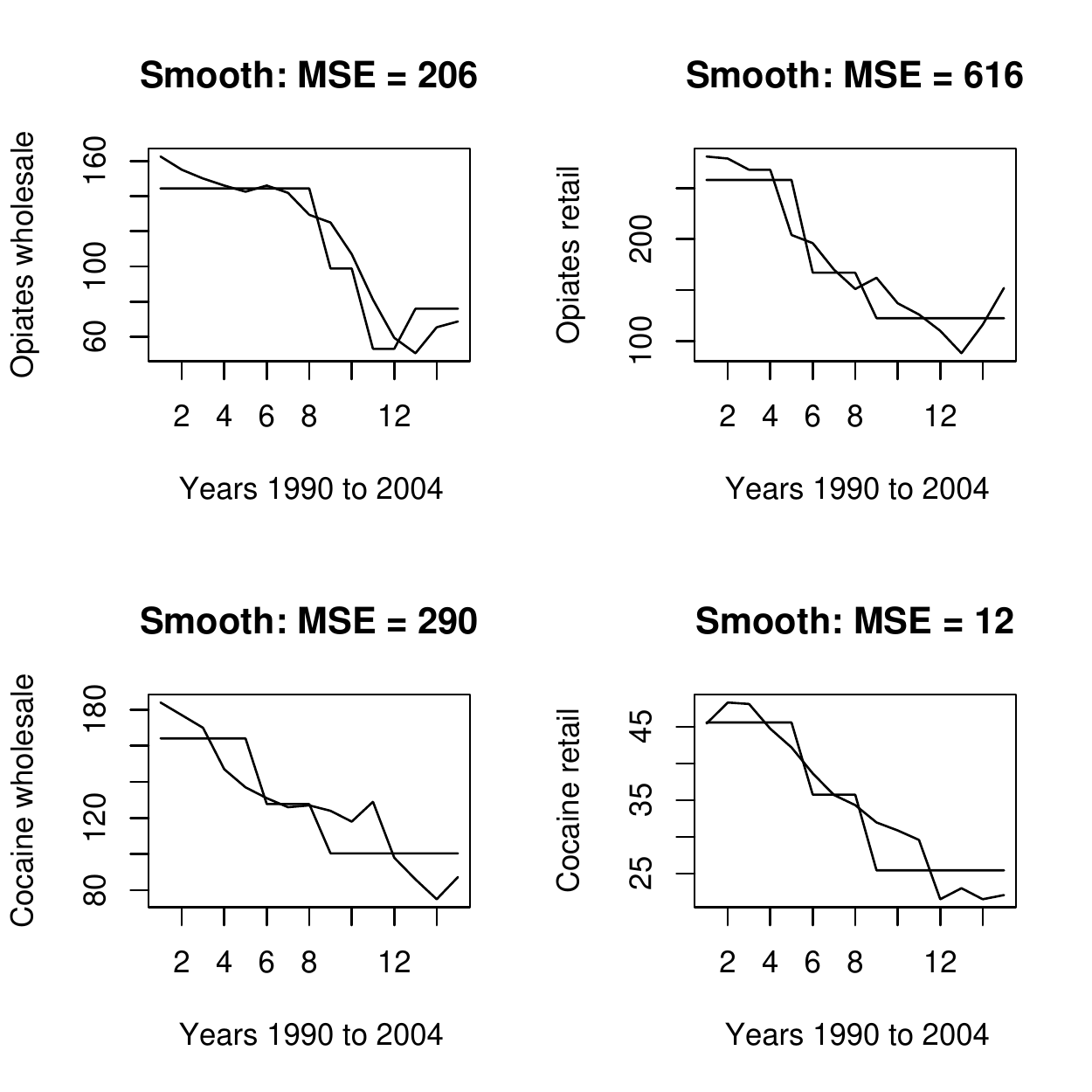}
\end{center}
\caption{Shown are the input signals with one selected smooth from,
respectively, Figures \protect\ref{fig5}, \protect\ref{fig6}, \protect\ref%
{fig7} and \protect\ref{fig8}. The same plots are shown separately in
Figures \protect\ref{fig9} and \protect\ref{fig10}. The smooth referred to
here is the piecewise linear smooth. The goodness of fit between piecewise
linear (more strictly, clusterwise constant) fit and original signal is
given by MSE, mean square error.}
\label{fig11}
\end{figure}

\subsection{Interpretation of Trends and Patterns} 

Figure \ref{fig11} shows panels containing (respectively from upper left to
lower right) opiates wholesale price and retail price signals; and cocaine
wholesale and retail price signals. Over these have been superimposed their
approximations that approximate the hierarchy, $\mathcal{H}$.

Note how for both wholesale signals, opiates and cocaine (left up and left
down panels in Figure \ref{fig11}), the hierarchy-representing signal is to
the left of or equal to the wholesale price signal from, and including,
years 8 = 1997 to 12 = 2001. In those years the information from the
hierarchy, expressing the Colombian conflict, was lower than the wholesale
price. Now, the wholesale price itself was falling. So we conclude that this
points to the Colombian internal conflict as having possibly played a
causative role in the wholesale price movement.

The corresponding situation for the retail price signals (right up and right
down in Figure \ref{fig11}) is different in this regard. The analogous
influence by the Colombian internal conflict on the retail price movement is
taking place from year 6 = 1995 to year 7 = 1996, and again from year 8 =
1997 to year 11 = 2000.

From a visual point of view there is a tighter fit between the opiates (top
two panels in Figure \ref{fig11}) compared to cocaine (bottom two panels in
Figure \ref{fig11}). Note that the opiates wholesale hierarchy fit is the
only one with 4 segments; all others have 3 segments. This was due to our
selection of partition. It is interesting to note the similarities of our
findings, i.e.\ the same years being indicated for the two wholesale
signals, and again the same years being indicated for the two retail signals.

Obviously we are dealing with one possible mechanism relating to price
formation here and there are other causal factors.

While we use 144 attributes in this work for tracking overall violence
levels, and while we find that breakpoints in the conflict occur at
consistent times relative to big movements in narcotics prices, we note
that these narcotics prices are largely falling in the period considered.
So it is not feasible to seek a link between prices and increasing 
or decreasing violence.  Ultimately we are looking for more subtle 
associations.  

An unanticipated finding that comes out of this analysis is as 
follows.  One could well expect the Colombian conflict to be a driver
of US cocaine prices.  What we have found is that the Colombian conflict
could also be influential in regard to US opiate prices.  We note this
as a potentially useful lead. 

\subsection{Novelties in Methodology}

Figure \ref{fig11} shows inter-related breakpoints arising from the
hierarchical interpretation of the Colombian conflict violence from 1990 to
2004. Each wavelet filtering and reconstructing of the data provides one set
of breakpoints. The simplest way to approximate any other signal -- for
example, the annually averaged wholesale price of opiates in the US -- is to
use a clusterwise constant fit to the signal \emph{with the breakpoints
taken into account}. This is what is done in Figure \ref{fig8888}. What is
innovative about the wavelet-based smoothing is how the breakpoints,
defining the piecewise linear approximation, are defined. They are defined
from the \emph{partial order} specified by $\mathcal{H}$.

\section{Conclusion}

We have described how hierarchy expresses changes at varying scales; how
hierarchy can be induced on a time-varying data set; and then how other data
sets defined on the same timeline can be ``folded over'' the hierarchically
structured data. Our aim is initial screening of relationships which may
lead to later stages of modeling, as exemplified by the following.

Taking the important products, coffee and oil, for the Colombian economy,
with the former being labor-intensive and the latter capital-intensive, \cite%
{dube} studied their relationship on the Colombian conflict at a low (fine
granularity) level: ``Exploiting exogenous price shocks in international
markets, we find that a drop in the price of coffee during the late 1990s
increased civil war violence in municipalities that cultivate coffee more
intensively. In contrast, a rise in oil prices during the same period
induced greater conflict in the oil municipalities relative to the non-oil
region. Lower coffee prices exacerbated conflict by reducing wages and the
opportunity cost of recruiting fighters into armed groups, while higher oil
prices fuelled conflict by increasing contestable revenue in local
governments, which invited predation by groups that steal these resources.''

Compared to this we have sought to preserve a wide range of levels of
aggregation. We do this because of our view that causative factors are often
on a range of resolution levels. Scale, in this case temporal, is important.
So rather than the simple smoothing of Figure \ref{fig8888} we instead
wanted to take more of the hierarchical information into account. Apart from
the goodness of fit being improved there were also a few intriguing findings
-- to be further studied with more detailed data -- relating to the
relationship between market processes and conflict violence.

\section*{Appendix 1: Attributes}

144 attributes were used for all events. Organisational acronyms used here:
FARC, Armed Revolutionary Forces of Colombia; ELN, National Liberation Army;
ERP, Popular Revolutionary Army; CGSB, Simon Bolivar Guerrilla Coordination;
DAS, Administrative Security Department; M-19, 19th of April Movement.

{\footnotesize K0, Number of civilians killed }

{\footnotesize K1, Number of military forces members killed }

{\footnotesize K2, Number of police members killed }

{\footnotesize K3, Number of paramilitaries killed }

{\footnotesize K4, Number of ELN members killed }

{\footnotesize K5, Number of FARC members killed }

{\footnotesize K6, Number of EPL members killed }

{\footnotesize K7, Number of ERP members killed }

{\footnotesize K8, Number of other guerrilla group members killed }

{\footnotesize K9, Number of other group members killed }

{\footnotesize K10, Number of other non-identified group members killed }

{\footnotesize K11, Number of CGSB members killed }

{\footnotesize K12, Number of DAS members killed }

{\footnotesize K13, Number of M19 members killed }

{\footnotesize I0, Number of civilians injured }

{\footnotesize I1, Number of military forces members injured }

{\footnotesize I2, Number of police members injured }

{\footnotesize I3, Number of paramilitaries injured }

{\footnotesize I4, Number of ELN members injured }

{\footnotesize I5, Number of FARC members injured }

{\footnotesize I6, Number of EPL members injured }

{\footnotesize I7, Number of ERP members injured }

{\footnotesize I8, Number of other guerrilla group members injured }

{\footnotesize I9, Number of other group members injured }

{\footnotesize I10, Number of other non-identified group members injured }

{\footnotesize I11, Number of CGSB members injured }

{\footnotesize I12, Number of DAS members injured }

{\footnotesize I13, Number of M19 members injured }

{\footnotesize KGue, Number of guerrilla members killed }

{\footnotesize KPar, Number of paramilitary members killed }

{\footnotesize KGob, Number of government members killed }

{\footnotesize KCiv, Number of civilians killed }

{\footnotesize K, Number of killings }

{\footnotesize IGue, Number of guerrilla members injured }

{\footnotesize IPar, Number of paramilitary members injured }

{\footnotesize IGob, Number of government members injured }

{\footnotesize ICiv, Number of civilians injured }

{\footnotesize I, Number of injured }

{\footnotesize KIGue, Number of guerrilla members casualties }

{\footnotesize KIPar, Number of paramilitary members casualties }

{\footnotesize KIGob, Number of government members casualties }

{\footnotesize KICiv, Number of civilians casualties }

{\footnotesize KI, Number of casualties }

{\footnotesize KClAtCivGue, Number of civilians killed in guerrilla events }

{\footnotesize KClAtCivPar, Number of civilians killed in paramiltary events 
}

{\footnotesize KClAtCivGob, Number of civilians killed in government forces
events }

{\footnotesize KICivELN, Number of civilian casualties in ELN events }

{\footnotesize KICivFARC, Number of civilian casualties in FARC events }

{\footnotesize KCivGueMas, Number of civilians killed in guerrilla massacres 
}

{\footnotesize KCivGueInc, Number of civilians killed in guerrilla
incursions }

{\footnotesize KCivGueBom, Number of civilians killed in guerrilla bombings }

{\footnotesize KGueGue, Number of guerrillas killed in guerrilla unilateral
attacks }

{\footnotesize KParGue, Number of paramilitaries killed in guerrilla
unilateral attacks }

{\footnotesize KGobGue, Number of government members killed in guerrilla
unilateral attacks }

{\footnotesize KCivGue, Number of civilians killed in guerrilla unilateral
attacks }

{\footnotesize KTotGue, Total number of killings in guerrilla unilateral
attacks }

{\footnotesize IGueGue, Number of guerrillas injured in guerrilla unilateral
attacks }

{\footnotesize IParGue, Number of paramilitaries injured in guerrilla
unilateral attacks }

{\footnotesize IGobGue, Number of government members injured in guerrilla
unilateral attacks }

{\footnotesize ICivGue, Number of civilians injured in guerrilla unilateral
attacks }

{\footnotesize ITotGue, Total number of injured in guerrilla unilateral
attacks }

{\footnotesize KIGueGue, Number of guerrilla casualties in guerrilla
unilateral attacks }

{\footnotesize KIParGue, Number of paramilitary casualties in guerrilla
unilateral attacks }

{\footnotesize KIGobGue, Number of government member casualties in guerrilla
unilateral attacks }

{\footnotesize KICivGue, Number of civilian casualties in guerrilla
unilateral attacks }

{\footnotesize KITotGue, Total number of casualties in guerrilla unilateral
attacks }

{\footnotesize KGuePar, Number of guerrillas killed in paramilitary
unilateral attacks }

{\footnotesize KParPar, Number of paramilitaries killed in paramilitary
unilateral attacks }

{\footnotesize KGobPar, Number of government members killed in paramilitary
unilateral attacks }

{\footnotesize KCivPar, Number of civilians killed in paramilitary
unilateral attacks }

{\footnotesize KTotPar, Total number of killings in paramilitary unilateral
attacks }

{\footnotesize IGuePar, Number of guerrillas injured in paramilitary
unilateral attacks }

{\footnotesize IParPar, Number of paramilitaries injured in paramilitary
unilateral attacks }

{\footnotesize IGobPar, Number of government members injured in paramilitary
unilateral attacks }

{\footnotesize ICivPar, Number of civilians injured in paramilitary
unilateral attacks }

{\footnotesize ITotPar, Total number of injured in paramilitary unilateral
attacks }

{\footnotesize KIGuePar, Number of guerrilla casualties in paramilitary
unilateral attacks }

{\footnotesize KIParPar, Number of paramilitary casualties in paramilitary
unilateral attacks }

{\footnotesize KIGobPar, Number of government member casualties in
paramilitary unilateral attacks }

{\footnotesize KICivPar, Number of civilian casualties in paramilitary
unilateral attacks }

{\footnotesize KITotPar, Total number of casualties in paramilitary
unilateral attacks }

{\footnotesize KGueGob, Number of guerrillas killed in government unilateral
attacks }

{\footnotesize KParGob, Number of paramilitaries killed in government
unilateral attacks }

{\footnotesize KGobGob, Number of government members killed in government
unilateral attacks }

{\footnotesize KCivGob, Number of civilians killed in government unilateral
attacks }

{\footnotesize KTotGob, Total number of killings in government unilateral
attacks }

{\footnotesize IGueGob, Number of guerrillas injured in government
unilateral attacks }

{\footnotesize IParGob, Number of paramilitaries injured in government
unilateral attacks }

{\footnotesize IGobGob, Number of government members injured in government
unilateral attacks }

{\footnotesize ICivGob, Number of civilians injured in government unilateral
attacks }

{\footnotesize ITotGob, Total number of injured in government unilateral
attacks }

{\footnotesize KIGueGob, Number of guerrilla casualties in government
unilateral attacks }

{\footnotesize KIParGob, Number of paramilitary casualties in government
unilateral attacks }

{\footnotesize KIGobGob, Number of government member casualties in
government unilateral attacks }

{\footnotesize KICivGob, Number of civilian casualties in government
unilateral attacks }

{\footnotesize KITotGob, Total number of casualties in government unilateral
attacks }

{\footnotesize KCivOth, Number of civilians killed in other group unilateral
attacks }

{\footnotesize ICivOth, Number of civilians injured in other group
unilateral attacks }

{\footnotesize KICivOth, Number of civilian casualties in other group
unilateral attacks }

{\footnotesize KGueClGobGue, Number of guerrillas killed in
government-guerrilla clashes }

{\footnotesize KParClGobGue, Number of paramilitaries killed in
government-guerrilla clashes }

{\footnotesize KGobClGobGue, Number of government members killed in
government-guerrilla clashes }

{\footnotesize KCivClGobGue, Number of civilians killed in
government-guerrilla clashes }

{\footnotesize KTotClGobGue, Total number of killings in
government-guerrilla clashes }

{\footnotesize IGueClGobGue, Number of guerrillas injured in
government-guerrilla clashes }

{\footnotesize IParClGobGue, Number of paramilitaries injured in
government-guerrilla clashes }

{\footnotesize IGobClGobGue, Number of government members injured in
government-guerrilla clashes }

{\footnotesize ICivClGobGue, Number of civilians injured in
government-guerrilla clashes }

{\footnotesize ITotClGobGue, Total number of injured in government-guerrilla
clashes }

{\footnotesize KIGueClGobGue, Number of guerrilla casualties in
government-guerrilla clashes }

{\footnotesize KIParClGobGue, Number of paramilitary casualties in
government-guerrilla clashes }

{\footnotesize KIGobClGobGue, Number of government member casualties in
government-guerrilla clashes }

{\footnotesize KICivClGobGue, Number of civilian casualties in
government-guerrilla clashes }

{\footnotesize KITotClGobGue, Total number of casualties in
government-guerrilla clashes }

{\footnotesize KGueClGobPar, Number of guerrillas killed in
government-paramilitary clashes }

{\footnotesize KParClGobPar, Number of paramilitaries killed in
government-paramilitary clashes }

{\footnotesize KGobClGobPar, Number of government members killed in
government-paramilitary clashes }

{\footnotesize KCivClGobPar, Number of civilians killed in
government-paramilitary clashes }

{\footnotesize KTotClGobPar, Total number of killings in
government-paramilitary clashes }

{\footnotesize IGueClGobPar, Number of guerrillas injured in
government-paramilitary clashes }

{\footnotesize IParClGobPar, Number of paramilitaries injured in
government-paramilitary clashes }

{\footnotesize IGobClGobPar, Number of government members injured in
government-paramilitary clashes }

{\footnotesize ICivClGobPar, Number of civilians injured in
government-paramilitary clashes }

{\footnotesize ITotClGobPar, Total number of injured in
government-paramilitary clashes }

{\footnotesize KIGueClGobPar, Number of guerrilla casualties in
government-paramilitary clashes }

{\footnotesize KIParClGobPar, Number of paramilitary casualties in
government-paramilitary clashes }

{\footnotesize IGobClGobPar, Number of government members injured in
government-paramilitary clashes }

{\footnotesize ICivClGobPar, Number of civilians injured in
government-paramilitary clashes }

{\footnotesize ITotClGobPar, Total number of injured in
government-paramilitary clashes }

{\footnotesize KIGueClGobPar, Number of guerrilla casualties in
government-paramilitary clashes }

{\footnotesize KIParClGobPar, Number of paramilitary casualties in
government-paramilitary clashes }

{\footnotesize KIGobClGobPar, Number of government casualties in
government-paramilitary clashes }

{\footnotesize KICivClGobPar, Number of civilian casualties in
government-paramilitary clashes }

{\footnotesize KITotClGobPar, Total number of casualties in
government-paramilitary clashes }

{\footnotesize KGueClGuePar, Number of guerrillas killed in
guerrilla-paramilitary clashes }

{\footnotesize KParClGuePar, Number of paramilitaries killed in
guerrilla-paramilitary clashes }

{\footnotesize KGobClGuePar, Number of government members killed in
guerrilla-paramilitary clashes }

{\footnotesize KCivClGuePar, Number of civilians killed in
guerrilla-paramilitary clashes }

{\footnotesize KTotClGuePar, Total number of killings in
guerrilla-paramilitary clashes }

{\footnotesize IGueClGuePar, Number of guerrillas injured in
guerrilla-paramilitary clashes }

{\footnotesize IParClGuePar, Number of paramilitaries injured in
guerrilla-paramilitary clashes }

{\footnotesize IGobClGuePar, Number of government members injured in
guerrilla-paramilitary clashes }

{\footnotesize ICivClGuePar, Number of civilians injured in
guerrilla-paramilitary clashes }

{\footnotesize ITotClGuePar, Total number of injured in
guerrilla-paramilitary clashes }

{\footnotesize KIGueClGuePar, Number of guerrilla casualties in
guerrilla-paramilitary clashes }

{\footnotesize KIParClGuePar, Number of paramilitary casualties in
guerrilla-paramilitary clashes }

{\footnotesize KIGobClGuePar, Number of government member casualties in
guerrilla-paramilitary clashes }

{\footnotesize KICivClGuePar, Number of civilian casualties in
guerrilla-paramilitary clashes }

{\footnotesize KITotClGuePar, Total number of casualties in
guerrilla-paramilitary clashes }

\section*{Appendix 2: the Correspondence Analysis Platform}

This Appendix and the next introduce important aspects of Correspondence
Analysis and hierarchical clustering. Further reading is to be found in \cite%
{benz} and \cite{murtaghca}.

\subsection*{Analysis Chain}

\begin{enumerate}
\item Starting point: a matrix that cross-tabulates the dependencies, e.g.\
frequencies of joint occurrence, of an observations crossed by attributes
matrix.

\item By endowing the cross-tabulation matrix with the $\chi^2$ metric on
both observation set (rows) and attribute set (columns), we can map
observations and attributes into the same space, endowed with the Euclidean
metric.

\item A hierarchical clustering is induced on the Euclidean space, the
factor space.

\item Interpretation is through projections of observations, attributes or
clusters onto factors. The factors are ordered by decreasing importance.
\end{enumerate}

Various aspects of Correspondence Analysis follow on from this, such as
Multiple Correspondence Analysis, different ways that one can encode input
data, and mutual description of clusters in terms of factors and vice versa.
In the following we use elements of the Einstein tensor notation of \cite%
{benz}. This often reduces to common vector notation.

\subsection*{Correspondence Analysis: Mapping $\protect\chi^2$ Distances
into Euclidean Distances}

\begin{itemize}
\item The given contingency table (or numbers of occurrence) data is denoted 
$k_{IJ} = \{ k_{IJ}(i,j) = k(i, j) ; i \in I, j \in J \}$.

\item $I$ is the set of observation indexes, and $J$ is the set of attribute
indexes. We have $k(i) = \sum_{j \in J} k(i, j)$. Analogously $k(j)$ is
defined, and $k = \sum_{i \in I, j \in J} k(i,j)$.

\item Relative frequencies: $f_{IJ} = \{ f_{ij} = k(i,j)/k ; i \in I, j \in
J\} \subset \mathbb{R}_{I \times J}$, similarly $f_I$ is defined as $\{f_i =
k(i)/k ; i \in I, j \in J\} \subset \mathbb{R}_I$, and $f_J$ analogously.

\item The conditional distribution of $f_J$ knowing $i \in I$, also termed
the $j$th \emph{profile} with coordinates indexed by the elements of $I$, is:

\[
f^i_J = \{ f^i_j = f_{ij}/f_i = (k_{ij}/k)/(k_i/k) ; f_i > 0 ; j \in J \}
\]
and likewise for $f^j_I$.

\item What is discussed in terms of information focusing in the text is
underpinned by the \emph{principle of distributional equivalence}. This
means that if two or more profiles are aggregated by simple element-wise
summation, then the $\chi^2$ distances relating to other profiles are not
effected.
\end{itemize}

\subsection*{Input: Cloud of Points Endowed with the Chi Squared Metric}

\begin{itemize}
\item The cloud of points consists of the couples: (multidimensional)
profile coordinate and (scalar) mass. We have $N_J(I) = $ $\{ ( f^i_J, f_i )
; i \in I \} \subset \mathbb{R}_J $, and again similarly for $N_I(J)$.

\item Included in this expression is the fact that the cloud of
observations, $N_J(I)$, is a subset of the real space of dimensionality $| J
|$ where $| . |$ denotes cardinality of the attribute set, $J$.

\item The overall inertia is as follows: 
\[
M^2(N_J(I)) = M^2(N_I(J)) = \| f_{IJ} - f_I f_J \|^2_{f_I f_J} 
\]
\[
= \sum_{i \in I, j \in J} (f_{ij} - f_i f_j)^2 / f_i f_j 
\]

\item The term $\| f_{IJ} - f_I f_J \|^2_{f_I f_J}$ is the $\chi^2$ metric
between the probability distribution $f_{IJ}$ and the product of marginal
distributions $f_I f_J$, with as center of the metric the product $f_I f_J$.

\item Decomposing the moment of inertia of the cloud $N_J(I)$ -- or of $%
N_I(J)$ since both analyses are inherently related -- furnishes the
principal axes of inertia, defined from a singular value decomposition.
\end{itemize}

\subsection*{Output: Cloud of Points Endowed with the Euclidean Metric in
Factor Space}

\begin{itemize}
\item The $\chi^2$ distance with center $f_J$ between observations $i$ and $%
i^{\prime }$ is written as follows in two different notations:

\[
d(i,i^{\prime i}_J - f^{i^{\prime }}_J \|^2_{f_J} = \sum_j \frac{1}{f_j}
\left( \frac{f_{ij}}{f_i} - \frac{f_{i^{\prime }j}}{f_{i^{\prime }}}
\right)^2 
\]

\item In the factor space this pairwise distance is identical. The
coordinate system and the metric change.

\item For factors indexed by $\alpha$ and for total dimensionality $N$ ($N = %
\mbox{ min } \{ |I| - 1, |J| - 1 \}$; the subtraction of 1 is since the $%
\chi^2$ distance is centered and hence there is a linear dependency which
reduces the inherent dimensionality by 1) we have the projection of
observation $i$ on the $\alpha$th factor, $F_\alpha$, given by $F_\alpha(i)$:

\begin{equation}
d(i,i^{\prime }) = \sum_{\alpha = 1..N} \left( F_\alpha(i) -
F_\alpha(i^{\prime }) \right)^2
\end{equation}

\item In Correspondence Analysis the factors are ordered by decreasing
moments of inertia.

\item The factors are closely related, mathematically, in the decomposition
of the overall cloud, $N_J(I)$ and $N_I(J)$, inertias.

\item The eigenvalues associated with the factors, identically in the space
of observations indexed by set $I$, and in the space of attributes indexed
by set $J$, are given by the eigenvalues associated with the decomposition
of the inertia.

\item The decomposition of the inertia is a principal axis decomposition,
which is arrived at through a singular value decomposition.
\end{itemize}

\subsection*{Dual Spaces and Transition Formulas}

\begin{itemize}
\item 
\[
F_\alpha(i) = \lambda^{-\frac{1}{2}}_\alpha \sum_{j \in J} f^i_j G_\alpha(j) %
\mbox{  for  } \alpha = 1, 2, \dots N; i \in I
\]

\[
G_\alpha(j) = \lambda^{-\frac{1}{2}}_\alpha \sum_{i \in I} f^j_i F_\alpha(i) %
\mbox{  for  } \alpha = 1, 2, \dots N; j \in J
\]

\item \emph{Transition formulas}: The coordinate of element $i \in I$ is the
barycenter of the coordinates of the elements $j \in J$, with associated
masses of value given by the coordinates of $f^i_j$ of the profile $f^i_J$.
This is all to within the $\lambda^{-\frac{1}{2}}_\alpha$ constant.

\item In the output display, the barycentric principle comes into play: this
allows us to simultaneously view and interpret observations and attributes.
\end{itemize}

\section*{Appendix 3: Hierarchical Clustering}

\subsection*{Sequence-Constrained Hierarchical Clustering}

Background on hierarchical clustering in general, and the particular
algorithm used here, can be found in \cite{murtagh85}. The sequence
constraint considered here is, for example, the total order involved in a
time series.

Consider the projection of observation $i$ onto the set of all factors
indexed by $\alpha$, $\{ F_\alpha(i) \}$ for all $\alpha$, which defines the
observation $i$ in the new coordinate frame. This new factor space is
endowed with the (unweighted) Euclidean distance, $d$. We seek a
hierarchical clustering that takes into account the observation sequence,
i.e.\ observation $i$ precedes observation $i^{\prime }$ for all $i,
i^{\prime }\in I$. We use the linear order on the observation set. \smallskip

\noindent Agglomerative hierarchical clustering algorithm:

\begin{enumerate}
\item Consider each observation in the sequence as constituting a singleton
cluster. Determine the closest pair of adjacent observations, and define a
cluster from them.

\item Determine and merge the closest pair of adjacent clusters, $c_1$ and $%
c_2$, where closeness is defined by $d(c_1, c_2) = \mbox{ max } \{
d_{ii^{\prime }}$ $\mbox{ such that } i \in c_1, i^{\prime }\in c_2 \}$.

\item Repeat the second step until only one cluster remains.
\end{enumerate}

This is a sequence-constrained complete link agglomeration criterion. The
cluster proximity at each agglomeration is strictly non-decreasing.

Recent application of this method can be found in \cite{murtaghpr}, relating
to the sequence of scenes in a film script (further discussed in 
\cite{merali}).

\subsection*{How a Hierarchy Expresses Change}

\begin{figure}[t]
\begin{center}
\includegraphics[width=6cm]{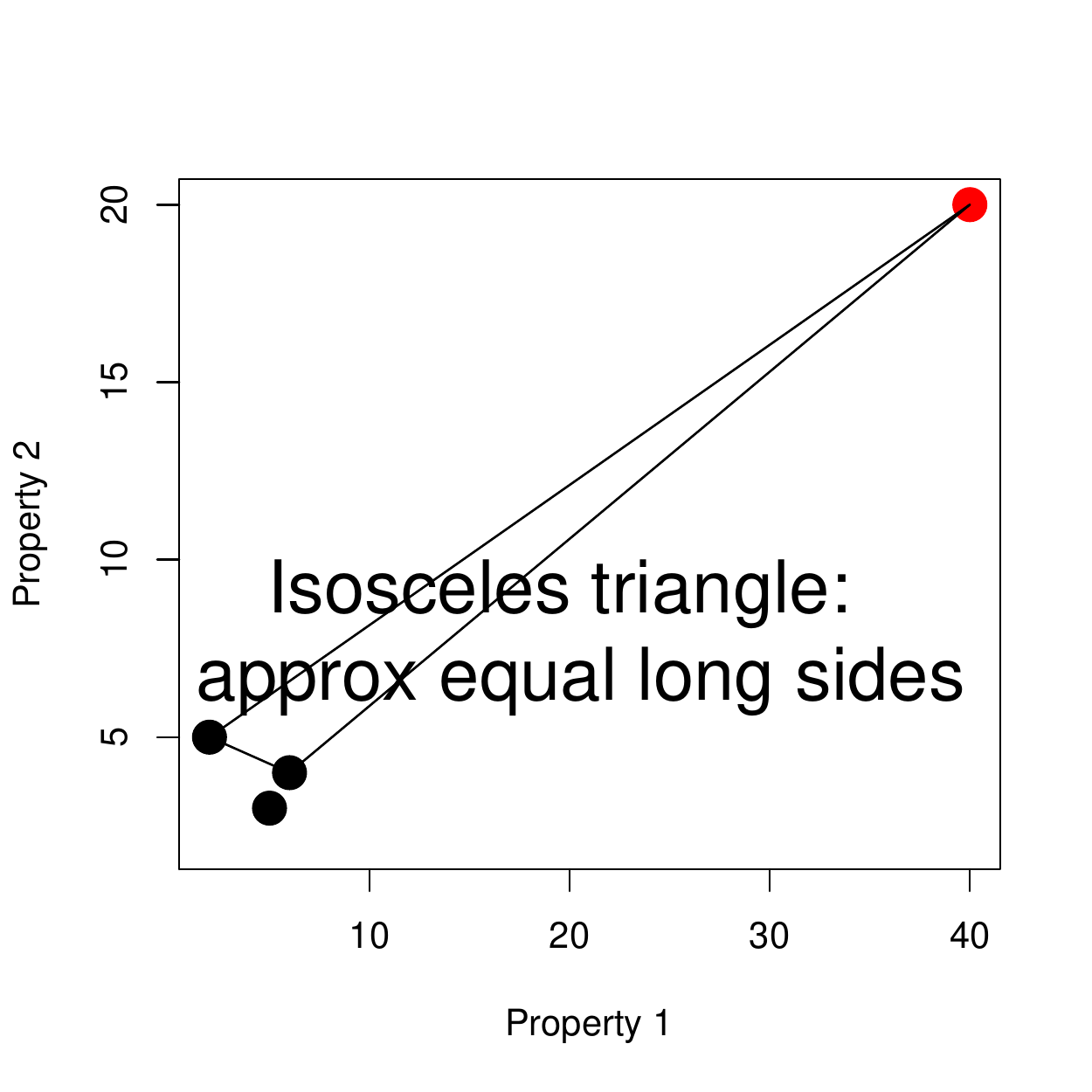} %
\includegraphics[width=6cm]{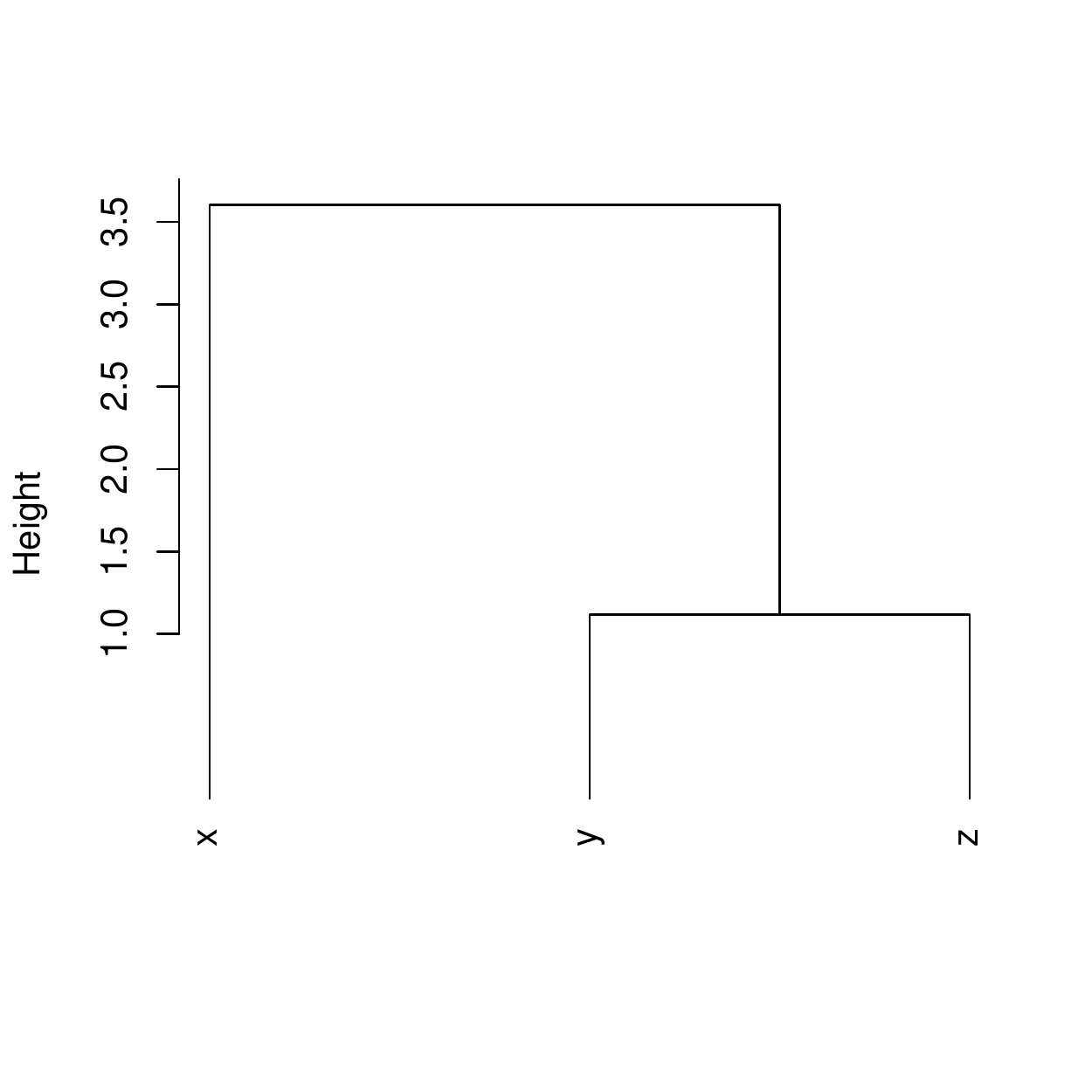}
\end{center}
\caption{Left: The query is on the far right. While we can easily determine
the closest target (among the three objects represented by the dots on the
left), is the closest really that much different from the alternatives?
Right: The strong triangular inequality defines an ultrametric: every
triplet of points satisfies the relationship: $d(d,z) \leq \mbox{max} \{
d(x,y), d(y,z) \}$ for distance $d$. Cf.\ by reading off the hierarchy, how
this is verified for all $x, y, z$: $d(x,z) = 3.5; d(x,y) = 3.5; d(y,z) = 1.0
$. In addition the symmetry and positive definiteness conditions hold for
any pair of points.}
\label{fig0}
\end{figure}

Consider Figure \ref{fig0}. The schematic representation, Figure \ref{fig0}
left, is of document retrieval where the query is in the upper right. In
Figure \ref{fig0} right the ultrametric distance is illustrated.

Further discussion of how a hierarchy expresses the semantics of change and
distinction, themes that are central in this article, can be found in \cite%
{murboole}.

The hierarchies that we induce on given data are based on an embedded set of
clusters. Through a series of $n - 1$ agglomerations starting from $n$
terminal nodes, the hierarchical clustering is constructed. The dendrogram
tree has, by construction, two-way branchings at each node. 

\section*{Appendix 4: Haar Wavelet Transform of a Dendrogram}

The discrete wavelet transform is a decomposition of data into spatial and
frequency components. In terms of a dendrogram these components are with
respect to, respectively, within and between clusters of successive
partitions. We show how this works taking the data of Table \ref{table5}.

\begin{table}[tbp]
\begin{center}
\begin{tabular}{|rrrrr|}
\hline
& Sepal.L & Sepal.W & Petal.L & Petal.W \\ \hline
1 & 5.1 & 3.5 & 1.4 & 0.2 \\ 
2 & 4.9 & 3.0 & 1.4 & 0.2 \\ 
3 & 4.7 & 3.2 & 1.3 & 0.2 \\ 
4 & 4.6 & 3.1 & 1.5 & 0.2 \\ 
5 & 5.0 & 3.6 & 1.4 & 0.2 \\ 
6 & 5.4 & 3.9 & 1.7 & 0.4 \\ 
7 & 4.6 & 3.4 & 1.4 & 0.3 \\ 
8 & 5.0 & 3.4 & 1.5 & 0.2 \\ \hline
\end{tabular}%
\end{center}
\caption{First 8 observations of Fisher's iris data. L and W refer to length
and width.}
\label{table5}
\end{table}

\begin{figure*}[tbp]
\begin{center}
\includegraphics[width=14cm]{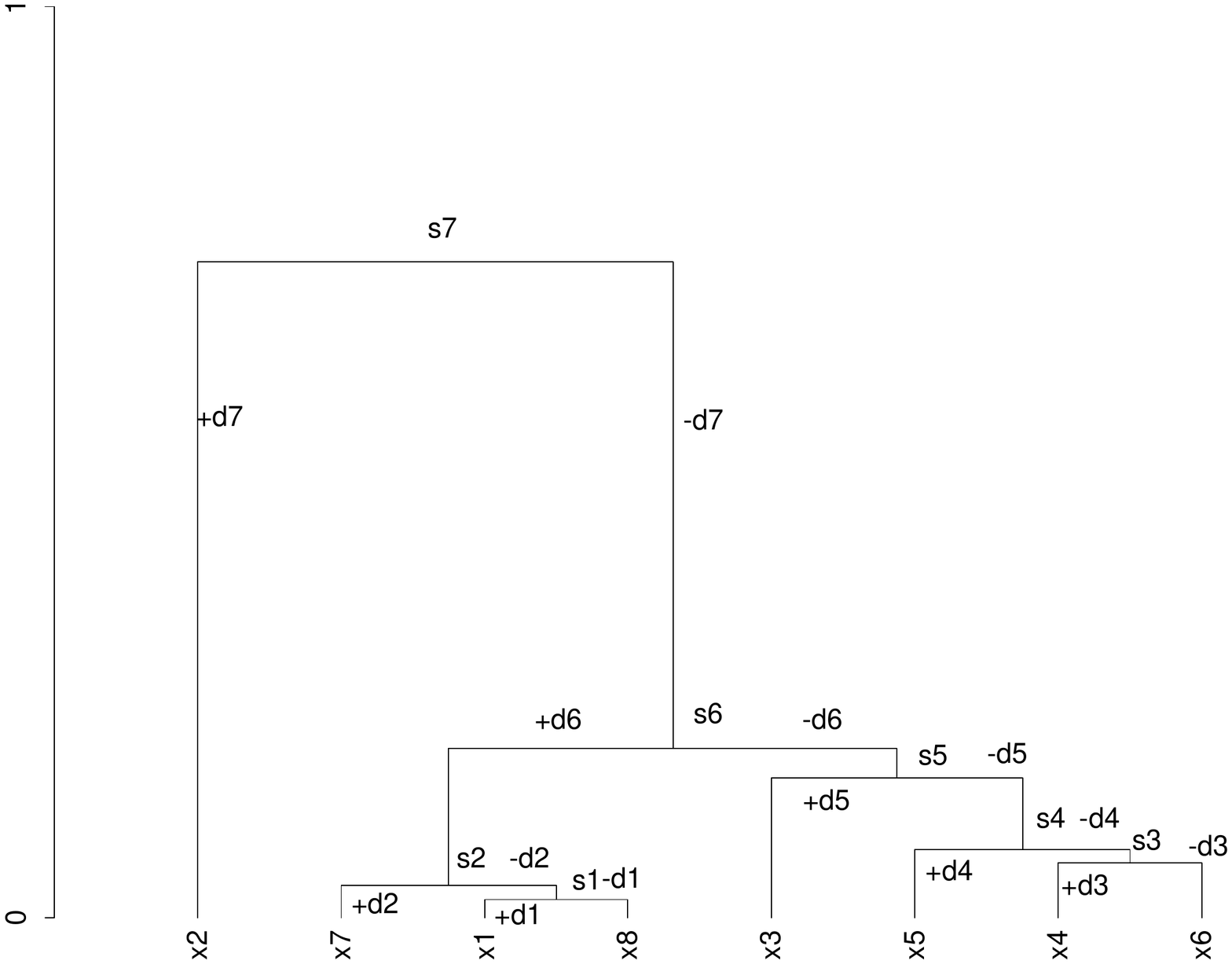}
\end{center}
\caption{Dendrogram on 8 terminal nodes constructed from first 8 values of
Fisher iris data. (Median agglomerative method used in this case.) Detail or
wavelet coefficients are denoted by $d$, and data smooths are denoted by $s$%
. The observation vectors are denoted by $x$ and are associated with the
terminal nodes. Each \emph{signal smooth}, $s$, is a vector. The (positive
or negative) \emph{detail signals}, $d$, are also vectors. All these vectors
are of the same dimensionality.}
\label{fig555}
\end{figure*}

The hierarchy built on the 8 observations of Table \ref{table5} is shown in
Figure \ref{fig555}.

Something more is shown in Figure \ref{fig555}, namely the detail signals
(denoted $\pm d$) and overall smooth (denoted $s$), which are determined in
carrying out the wavelet transform, the so-called forward transform.

The inverse transform is then determined from Figure \ref{fig555} in the
following way. Consider the observation vector $x_2$. Then this vector is
reconstructed exactly by reading the tree from the root: $s_7 + d_7 = x_2$.
Similarly a path from root to terminal is used to reconstruct any other
observation. If $x_2$ is a vector of dimensionality $m$, then so also are $%
s_7$ and $d_7$, as well as all other detail signals.

\begin{table*}[tbp]
\begin{center}
\begin{tabular}{|rrrrrrrrr|}
\hline
& s7 & d7 & d6 & d5 & d4 & d3 & d2 & d1 \\ \hline
Sepal.L & 5.146875 & 0.253125 & 0.13125 & 0.1375 & $-0.025$ & 0.05 & $-0.025$
& 0.05 \\ 
Sepal.W & 3.603125 & 0.296875 & 0.16875 & $-0.1375$ & 0.125 & 0.05 & $-0.075$
& $-0.05$ \\ 
Petal.L & 1.562500 & 0.137500 & 0.02500 & 0.0000 & 0.000 & $-0.10$ & 0.050 & 
0.00 \\ 
Petal.W & 0.306250 & 0.093750 & $-0.01250$ & $-0.0250$ & 0.050 & 0.00 & 0.000
& 0.00 \\ \hline
\end{tabular}%
\end{center}
\caption{The hierarchical Haar wavelet transform resulting from use of the
first 8 observations of Fisher's iris data shown in Table \protect\ref%
{table5}. Wavelet coefficient levels are denoted d1 through d7, and the
continuum or smooth component is denoted s7.}
\label{table6}
\end{table*}

This procedure is the same as the Haar wavelet transform, only applied to
the dendrogram and using the input data.

A complete specification of this wavelet transform for the data in Table \ref%
{table5} is shown in Table \ref{table6}.

The principle of ``folding'' the hierarchy onto an external signal is as
follows. The wavelet transform codifies the hierarchy. Having that, we apply
the ``codification'' of the hierarchy with the new, external signal as input.

Wavelet regression entails setting small and hence unimportant detail
coefficients to 0 before applying the inverse wavelet transform.

More discussion can be found in \cite{murhaar}.

\end{document}